\RequirePackage{fix-cm}
\documentclass[smallcondensed]{svjour3}     
\smartqed  
\usepackage{graphicx}
\usepackage{amsmath}
\usepackage{amssymb}
\usepackage{array}
\usepackage{subfigure}
\usepackage{longtable}
\usepackage{multirow}
\usepackage{natbib}
\newcolumntype{C}[1]{>{\centering\let\newline\\\arraybackslash\hspace{0pt}}m{#1}}

\usepackage[dvipsnames]{xcolor}

\begin{document}

\title{Reinforcement Learning for Multi-Product Multi-Node Inventory Management in Supply Chains
}
\subtitle{Handling warehouses, trucks, and stores with finite capacity}

\titlerunning{RL for Multi-Product Multi-Node Inventory Management}        

\author{Nazneen~N~Sultana         \and
        Hardik~Meisheri           \and
        Vinita~Baniwal            \and
        Somjit~Nath               \and
        Balaraman~Ravindran       \and
        Harshad~Khadilkar
}

\authorrunning{N.N. Sultana et al.} 

\institute{N.~N.~Sultana, H.~Meisheri, V.~Baniwal, S.~Nath, H.~Khadilkar \at
              TCS Research, Mumbai, India \\
              (Contact author) \email{nn.sultana@tcs.com}           
           \and
           B.~Ravindran \at
              Robert Bosch Centre for Data Science and AI, IIT Madras, Chennai, India \\
             \email{ravi@cse.iitm.ac.in}
}

\date{Received: March 15, 2020 / Under review} 

\maketitle

\begin{abstract}
This paper describes the application of reinforcement learning (RL) to multi-product inventory management in supply chains. The problem description and solution are both adapted from a real-world business solution. The novelty of this problem with respect to supply chain literature is (i) we consider concurrent inventory management of a large number (50 to 1000) of products with shared capacity, (ii) we consider a multi-node supply chain consisting of a warehouse which supplies three stores, (iii) the warehouse, stores, and transportation from warehouse to stores have finite capacities, (iv) warehouse and store replenishment happen at different time scales and with realistic time lags, and (v) demand for products at the stores is stochastic. We describe a novel formulation in a multi-agent (hierarchical) reinforcement learning framework that can be used for parallelised decision-making, and use the advantage actor critic (A2C) algorithm with quantised action spaces to solve the problem. Experiments show that the proposed approach is able to handle a multi-objective reward comprised of maximising product sales and minimising wastage of perishable products.
\end{abstract}


\section{Introduction} \label{sec:intro}

A moderately large retail business may be composed of approximately 1,000 stores spread over a large geographical area, with each store selling up to 100,000 product types each. The inventory of products in each store is periodically replenished by trucks\footnote{For example, once per day or once every few hours.}. These trucks originate from a local warehouse, which serves a set of 10-20 stores within a smaller region (such as a city). A truck may visit more than one store, subject to time, volume, and weight constraints. Conversely, if the stores are large, more than one truck may serve the same store. The critical decision to be made in this step of the supply chain is the replenishment quantity of each product in every store, at every delivery time period. There are multiple tradeoffs involved in this process, including maintenance of inventory in the store, minimisation of wastage due to products going past their sell-by dates, fairness across the product range, and capacity sharing on the truck. Moving one step up the supply chain hierarchy, the warehouse is tasked with maintaining sufficient inventory of products for serving 10-20 stores for a given time duration. The replenishment delivery intervals for warehouses are typically 4x-7x longer than those for stores\footnote{If stores are replenished four times a day, the warehouse may be replenished once per day. If stores are replenished once per day, the warehouse may be replenished once per week.}. 

\begin{figure}[t]
	\centering
	\includegraphics[width=0.9\textwidth]{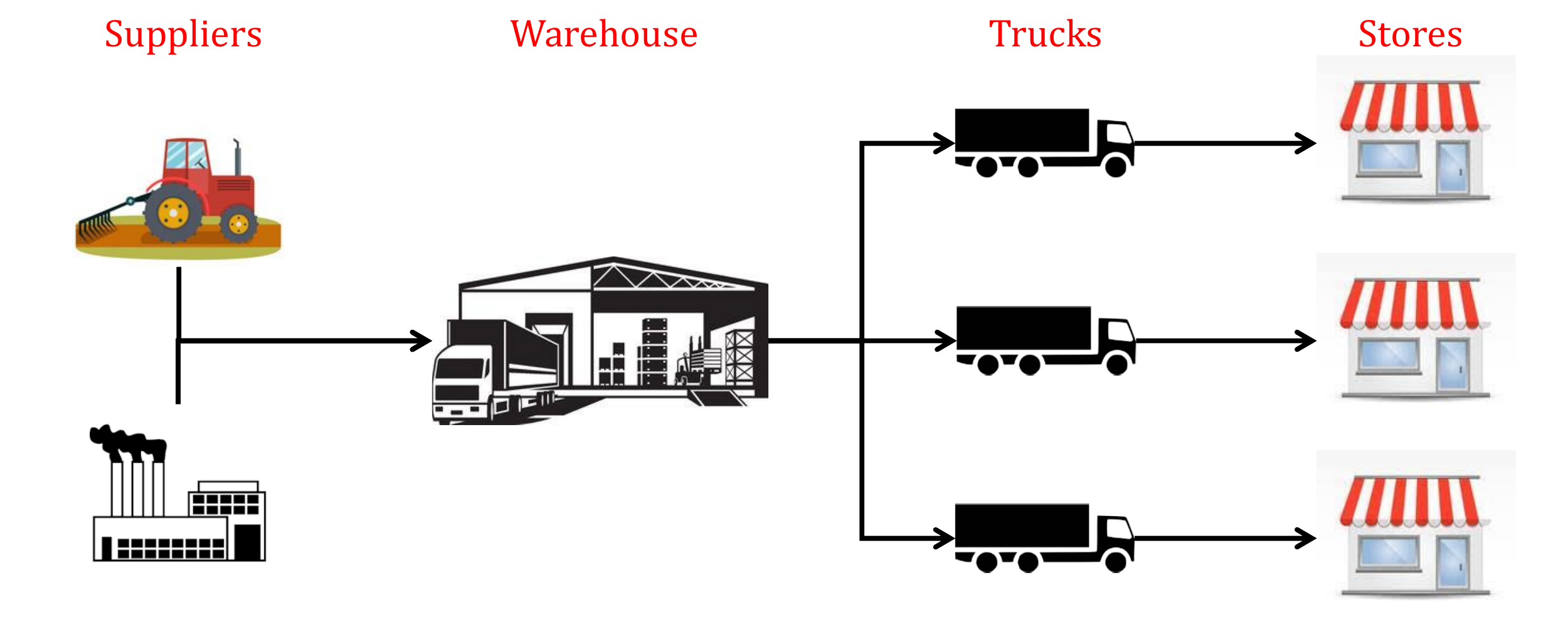}
	\caption{Flow of products within a retail supply chain. The goal is to maintain inventory of the full range of products in the stores, while minimising costs.}
	\label{fig:sc-eg}
\end{figure}

\textbf{Challenges: }
In our experience while working with retail businesses, the difficulty of keeping the replenishment system working smoothly can be broken down into three individual problems. First and foremost, retailers worry about the problem of \textbf{under-stocking}: running out of inventory in the stores, for one or more products. This can happen because of an unexpected surge in demand for some products (demand variability), delays or shortfalls from suppliers/manufacturers, or due to miscalculations when managing inventory within the internal supply chain. The second challenge is that of \textbf{over-stocking} or \textit{wastage}, which happens when too much inventory is stocked in the store. This not only causes inconvenience when the stores have insufficient backroom storage, but can also lead to losses of perishable products. Stocking inventory in a warehouse is meant to mitigate under- and over-stocking, but this is difficult because of the third challenge of \textbf{computational complexity}. If hundreds or thousands of products are competing for the same labour, transport, and shelf capacities, computing the correct replenishment quantities for each product is difficult. Furthermore, product attributes are highly heterogeneous, and affect the operations in different ways. Even when heuristic algorithms tuned for specific retail businesses over several years are able to manage the first three challenges, they frequently result in \textbf{high operating cost}. Given the conflicting objectives of high availability, low wastage, and low vendor purchase costs, heuristics tend to operate in a narrow region of states and actions. By contrast, learning based systems can explore a larger (and potentially better) portion of the state-action spaces, as explained below.

\textbf{Case for using learning-based replenishment systems: }
State-of-the-art in supply chain management contains several automation efforts [\cite{fernie2018logistics}], with varying degrees of success. We describe prior literature in detail in the Section \ref{sec:related}. The key takeaway from reported studies, is that they fall under one of two categories. First, practical applications typically automate individual tasks in the supply chain (for example, store replenishment) using heuristics or rule-based systems. This approach requires design of hand-tuned heuristics for each instance, since business goals can be substantially different from one retailer to another. Additionally, it is difficult to define heuristics for concurrently optimising the end-to-end operations in the supply chain. As a consequence, state-of-the-art practical systems tend to (sub-)optimise in silos without effective cooperation across supply chain nodes. 

A second set of studies use data-driven methods such as reinforcement learning for decisions across multiple nodes, but tend to focus on a single product. While this accounts for stochasticity in customer demand and models characteristics such as the bullwhip effect, it overlooks system constraints such as shared capacity, which make the multi-product problem orders of magnitude larger than the single-product problem. The combinatorial nature of the problem as a result of shared capacity across hundreds of products, and effective coordination between the warehouse and the stores, are one of our principal interests in this problem.

The motivation for choosing reinforcement learning (specifically, multi-agent RL) as the method of choice is summarised in the following list.
\begin{itemize}
    \item RL allows us to define a generic set of features (states) and decisions (actions), while learning the specific policies for individual problem instances on its own.
    \item Multi-agent RL offers a way to coordinate between stores and warehouses while maximising system goals.
    \item Each agent in the MARL framework operates and adapts on its own, with all other agents and external factors being absorbed in the environment. This allows us to deploy new agents into the supply chain without modifying any of the prior deployments. 
    \item Practically, this means we can add/remove stores or warehouses in the supply chain independently, and reasonably expect all other nodes to adjust their policies as needed. For fresh deployments, we can also choose to move decision-making systems from their current versions to RL one node at a time, reducing the risk of operational destabilisation.
    \item The RL state and action space definitions used in this work allow the algorithm to work with an arbitrary number of products, as well as an arbitrary number of stores to be supplied by a given warehouse.
\end{itemize}

In this paper, we are interested in inventory management for supply chains with a hierarchical structure of product flows. We consider a sample scenario with a single warehouse, which supplies three stores at periodic intervals. The stores are replenished every 6 hours (four times a day), and the warehouse is replenished every 24 hours (once a day). The hierarchical structure is imposed in Section \ref{sec:desc} by the assumption that (i) all products are delivered to stores only from the warehouse, (ii) there is no lateral flow of products from one store to another, and (iii) the transportation capacity for each store is independent of the other stores\footnote{Practically, this means that the deliveries happen on separate trucks.}. As the development in Section \ref{sec:method} shows, the algorithm is flexible enough to work with weaker constraints with minor modifications. The results in Section \ref{sec:results} show that the proposed methodology is scalable enough to work with a few tens of products (50) up to a much larger number of products (1000), without requiring extensive hyperparameter tuning from one instance to another.

	
	

\textbf{Contributions and paper structure: }
We believe that the principal contributions of this paper are (i) formulating a Multi-Agent RL approach to solve the multi-product constrained inventory management problem, (ii) achieving effective coordination between the warehouse and stores, through reward and state sharing and parallelized decisions within each agent, and (iii) experimentation and detailed evaluation of the learned policies in a realistic two-echelon (warehouse and stores) setting, on instances of small (50 products), medium (220 products), and large (1000 products) size.

The rest of this paper is structured as follows. Section \ref{sec:related} gives an overview of past literature from traditional methods to current RL driven methods for supply chain problems. A detailed formulation of the problem as a two-echelon supply chain inventory problem is presented, including local and global constraints, in Section \ref{sec:desc}. Section \ref{sec:method} deals with the proposed method in solving this problem with Deep Reinforcement Learning. Experiments on three different sets of data with a varying number of products (50, 220, and 1000 products) and evaluation of the learned policies are presented in Section \ref{sec:results}.


\section{Related work} \label{sec:related}

We classify prior literature into (i) traditional approaches for inventory management, (ii) data-driven approaches, and (iii) the use of RL in related problem areas. 

\textbf{Traditional models for inventory management: }
The formulation described in Section \ref{sec:desc} shows that the inventory management problem is related to the multivariable dynamical system control problem, involving a vector of (possibly interdependent) states. Examples of multivariable problems include physical dynamical systems such as the position and its derivatives of an inverted pendulum [\cite{pathak2005velocity}], as well as distributed systems such as transportation networks [\cite{de2010multi}] and inventory management in supply chains [\cite{van1997production}]. 

Most of the literature in the inventory management field takes one of two paths. One set considers multiple nodes in the supply chain [\cite{nahmias1994optimizing,Nahmias1993,lee1993material}], but for a single product. The focus of these studies is on mitigating well-known issues such as the bullwhip effect\footnote{The bullwhip effect occurs when a lag in demand forecasts causes growing oscillations in inventory levels, analogous to the motion of a whip.}. The second set of studies consider multiple products, but for a single node in the supply chain. Instances at relatively small scale are solved as joint assortment-stocking problems using MILP [\cite{smith2000management,caro2010inventory}] and related techniques such as branch-and-cut [\cite{coelho2014optimal}]. However, these techniques have large computation times and are limited to problems with a handful of product types (fewer than 10) and short time horizons. Implementations at practical scales typically operate with simple heuristics such as threshold-based policies [\cite{condea2012rfid}] or formulae based on demand assumptions [\cite{cachon1997campbell,silver1979simple}]. The takeaway is that traditional methods in inventory management do not appear to naturally and simultaneously handle a large number of products moving through multiple nodes of the supply chain. It is possible to use such approaches through post-processing (we use one such method as a baseline in Section \ref{sec:results}), but the results are suboptimal.

\textbf{Data-driven methods: }
A more promising approach is to tune the parameters of the controller or decision-making agent empirically, using historical or self-generated data. Adaptive control, Model-predictive control, Imitation learning, and Approximate dynamic programming are possible options. Adaptive control (AdC) [\cite{aastrom2013adaptive}] is suitable for physical dynamical systems, where the control law is defined as a functional relationship while the parameters are computed using empirical data. However, adaptive control typically requires analytical models of the control and adaptation laws. As a result, their use in inventory management [\cite{garcia2012inventory}] is limited to cases where the dynamics of the supply chain can be identified explicitly, and furthermore are invariant over time. 

Model predictive control (MPC) [\cite{camacho2013model}] is less restrictive, because it does not require an explicit control law. However, it does require rollouts over multiple time steps. The computational complexity and stochasticity introduced by this aspects limits the application of MPC to instances with a few products [\cite{braun2003model}]. Imitation learning (IL) [\cite{schaal1999imitation}] learns from expert behaviour, assuming that the ideal policy is able to maximise whatever objective is being targeted. The inherent problems of design complexity and performance limitations apply here as well; to the definition of the expert policy rather than to the IL algorithm. Additionally, the general form of the problem may not admit an obvious expert policy to train with. While this is feasible for instances where the optimal replenishment policy can be computed [\cite{baniwal2019imitation}], this is not possible in the present instance. In fact, all three methods (AdC, MPC, IL) require knowledge of ideal behaviour or trajectory of the system, which is not available in the high-dimensional, multi-node inventory management scenario.

Approximate Dynamic Programming (ADP) [\cite{bertsekas2005dynamic,powell2007approximate}] also depends on analytical forms of the value function, and requires explicit state transition probabilities and stage costs. However, some variants such as Adaptive Critics [\cite{shervais2000adaptive,shervais2003intelligent,si2004handbook}] and Adaptive Dynamic Programming [\cite{papadaki2003adaptive}] have been previously used for the inventory management problem (sometimes called the product dispatching problem), both with multi-product and multi-node flavours. There do not appear to be prior studies that consider both multi-product and multi-node characteristics simultaneously, especially when the number of products can be as high as hundreds or even thousands. Furthermore, the studies either use explicit value function forms [\cite{papadaki2003adaptive}], or a centralised critic [\cite{shervais2003intelligent}] where neural networks are employed. In this paper, our principal interest is in the multi-product, multi-node inventory management problem where the number of products as well as the number of nodes (stores) can change over time. Therefore, we focus on a distributed (multi-agent) RL approach that is agnostic to the number of products and nodes.

\textbf{RL techniques in inventory management and related areas: }
    Some prior studies use reinforcement learning for the inventory management problem [\cite{giannoccaro2002inventory,kara2018reinforcement,jiang2009case}], but principally for the single-product version. They also focus on specific goals, such as maximum profit [\cite{giannoccaro2002inventory}], minimum operating cost [\cite{kara2018reinforcement}], or maintaining target inventory [\cite{jiang2009case}]. Multi-product inventory management studies are so far limited to a single node [\cite{barat2019actor}]. In this work, we consider the inventory management problem from the point of view of (i) the warehouse, which orders inventory from vendors with a predefined lead time, and (ii) multiple stores, which request replenishment from a common warehouse. Furthermore, we consider problems where hundreds of products share transportation capacity on the way to the stores, and the warehouse inventory of each given product is shared by all the stores.

We take inspiration from multi-agent reinforcement learning [\cite{littman1994markov}] applied to related problems such as economic dispatch [\cite{mannion2016dynamic}] and transportation [\cite{el2013multiagent,mousavi2017traffic}]. We also use insights from work on reduction of continuous action space complexity through quantization in robotics [\cite{kober2013reinforcement,theodorou2010reinforcement}] and our own work in operations research [\cite{verma2019containers,khadilkar2019scalable,barat2019actor}]. The state and action space formulation described in Section \ref{sec:method} is simple enough for value-based RL methods such as Deep Q-Networks [\cite{mnih2015human,van2016deep}] to work. We do not incorporate recent work on action space reduction through branching DQN [\cite{tavakoli2018action}] or action embedding [\cite{chandak2019learning}], in order to keep the models small, easy to train, and explainable enough to be acceptable to business users.

While the topology of our network (warehouse and multiple stores) is hierarchical, the problem described in Section \ref{sec:desc} is not directly a hierarchical reinforcement learning problem [\cite{vezhnevets2017feudal,nachum2018data}]. This is because the warehouse and stores are not working towards a single objective, with the warehouse (high-level agent) setting goals for the stores (low-level agent). Instead, the stores are tasked with maintaining inventory levels of products (in addition to secondary goals), while the warehouse is tasked with ensuring sufficient inventory levels to supply demand from stores.

\section{Problem Description} \label{sec:desc}

Let us consider a compact version of the problem, by looking at one warehouse, a set of $S$ stores, and a set of discrete time steps $t$. 

\subsection{Store model}

Each store sells a range of $P$ products. The inventory of product $1\leq i \leq P$ in store $1\leq j \leq S$ at time step $t$ is denoted by $x_{ij}(t)$. At each time step, the warehouse supplies a quantity $u_{ij}(t)$ of product $i$, in order to replenish the inventory. The change in state is assumed to be instantaneous, and is given by,
\begin{equation}
    {\bf{x}}_j(t)^+ = {\bf{x}}_j(t)^- + {\bf{u}}_j(t), \label{eq:repldyn}
\end{equation}
where ${\bf{x}}_j(t)$ and ${\bf{u}}_j(t)$ are $P$ element vectors corresponding to all products in store $j$. The constraints on the system from the point of view of a store are of the form,
\begin{align}
{\bf{0}} \leq {\bf{x}}_j(t) & \leq {\bf{1}}, \label{eq:inv} \\
{\bf{0}} \leq {\bf{u}}_j(t) & \leq {\bf{1}}, \label{eq:control} \\
{\bf{0}} \leq {\bf{x}}_j(t)^- + {\bf{u}}_j(t) & \leq {\bf{1}}, \label{eq:shelf} \\
{\bf{v}}^T\,{\bf{u}}_j(t) & \leq v_{\mathrm{max},j}. \label{eq:truckvol}
\end{align}
Here, constraints (\ref{eq:inv}) and (\ref{eq:control}) are related to the range of acceptable values of each product (assuming normalisation of inventory for each product to the range $[0,1]$). Constraint (\ref{eq:shelf}) states that the level of inventory just after replenishment (${\bf{x}}_j(t)^+$ according to (\ref{eq:repldyn})) cannot exceed the maximum inventory level. Constraint (\ref{eq:truckvol}) sets the maximum value of the total replenishment quantity to $v_{\mathrm{max},j}$, mimicking transportation capacity limitations. Column vector $\bf{v}$ corresponds to the unit volume for each product $i$. Finally, we assume the existence of some estimator for sales during the current time step of the form,
\begin{equation}
    \hat{{\bf{w}}}_j(t+1) = {\bf{f}}\left( {\bf{w}}_j(0:t) \right),
    \label{eq:estimator}
\end{equation}
where ${\bf{w}}(0:t)$ is the amount of products sold within every time period from the beginning. The actual inventory after sales to customers is given by,
\begin{equation}
    {\bf{x}}_j(t+1)^- = \max \left( {\bf{0}}, {\bf{x}}_j(t)^+ - {\bf{w}}_j(t+1) \right),
    \label{eq:invchange}
\end{equation}
where ${\bf{w}}_j(t+1)$ is the true realised value of sales. Inventory management is a multi-objective optimisation problem, with direct costs relating to (i) the reduction of inventory for some products to 0, commonly known as out-of-stock, and (ii) the quantity $q_{\mathrm{waste},i}(t)$ of products wasted (spoiled) during the time period ending at $t$. In addition, we wish to ensure that some products are not unfairly preferred over others when the system is stressed (for example, when the capacity $v_\mathrm{max}$ is too small to keep up with product sales). Therefore, we include a fairness penalty on the variation in inventory levels across the product range, from the 95$^\mathrm{th}$ to the 5$^\mathrm{th}$ percentile in store $j$ (denoted by $\Delta {\bf{x}}_j(t)^\mathrm{.95}_{\mathrm{.05}}$) across all products. The cost incurred by each store $j$ during time interval $t$ is defined in (\ref{eq:invreward}). 
\begin{equation}
\mathcal{C}_{j,\mathrm{st}}(t) =  \underbrace{\frac{p_{j,\mathrm{empty}}(t)}{P}}_{\text{Out of stock}} + \underbrace{\frac{\sum_{i} q_{\mathrm{waste},ij}(t)}{P}}_{\text{Wastage}} + \underbrace{\Delta {\bf{x}}_j(t)^\mathrm{.95}_{\mathrm{.05}}}_{\text{Percentile spread}},\label{eq:invreward}
\end{equation} 
where $P$ is the total number of products, $p_\mathrm{empty}(t)$ is the number of products with $x_{ij}=0$ at the end of the previous time period. The selfish goal of each store is to minimise the discounted sum of this cost from the current time period onward.

\subsection{Warehouse model}

The warehouse is obliged to supply inventory ${\bf{u}}_j$ to store $j$ as long it holds sufficient inventory at that time. Therefore, an additional system constraint is imposed by inventory levels $\chi (t)$  in the warehouse,
\begin{equation}
    \sum_{j=1}^S a_j {\bf{u}}_j(t) \leq \chi (t),
    \label{eq:warehouse}
\end{equation}
where $a_j$ is a constant multiplier for converting the normalised inventories of each store to a common reference value, so that $0 \leq \chi_i (t) \leq 1$ for each product $i$. The replenishment of $\chi$ itself happens through a second decision-making exercise, after every $n$ time periods of store replenishment. Furthermore, the action (vendor orders) computed at time period $nt$ only gets implemented (products delivered) after $n$ time periods.
\begin{equation}
    \chi (n(t+1))^+ = \chi (n(t+1))^- + \mu (nt), \label{eq:replwarehouse}
\end{equation}
subject to the constraints,
\begin{align}
{\bf{0}} \leq \chi (n(t+1)) & \leq {\bf{1}}, \label{eq:wareinv} \\
{\bf{0}} \leq \mu (nt) & \leq {\bf{1}}, \label{eq:warecontrol} \\
{\bf{0}} \leq \chi (n(t+1))^- + \mu (nt) & \leq {\bf{1}} \label{eq:wareshelf}.
\end{align}
These constraints are direct counterparts of the physical store constraints (\ref{eq:inv})-(\ref{eq:shelf}), except that the quantity of products delivered to the warehouse is unconstrained. This is because the products are assumed to be provided directly by separate vendors. Similar to the estimator for sales in the stores in the next time period, we assume the existence of two estimators in the case of warehouse replenishment. Given a time instant $nt$, the first estimator predicts the warehouse inventory at time $n(t+1)$, which is when the chosen action $\mu(nt)$ will take effect. The second estimator predicts the aggregate replenishment demand from stores, during the $n$ time periods after that ($n(t+1)$ to $n(t+2)$).
\begin{align}
    \hat{\chi}(n(t+1)) & = {\bf{F_1}}(\chi(nt),{\bf{W}}(0:nt)), \text{ and } \nonumber \\
    \hat{{\bf{W}}}(n(t+2)) & = {\bf{F_2}}\left( {\bf{W}}(0:nt) \right),
    \label{eq:ware_estimator}
\end{align}
where ${\bf{W}}(0:nt)$ is the amount of replenishment demanded within every time period from the beginning. Instead of a volume constraint, the replenishment of the warehouse is subject to a cost given by,
\begin{equation}
    \mathcal{C}_\mathrm{wh}(nt) =  \frac{1}{P}\sum_{i=1}^P \left[ Q_{\mathrm{waste},i} + b_i(nt)\, \left( C_i + \mu_i(nt)\,\Tilde{C}_i \right) \right].
    \label{eq:ware_cost}
\end{equation}
The multiplier $b_i(nt)$ is a binary flag indicating whether product $i$ is being replenished in this period (non-zero $\mu_i$) or not. For products that are being ordered from vendors, there is a fixed cost $C_i$ and a variable cost $\Tilde{C}_i$ proportional to the quantity of replenishment\footnote{In real operations, the fixed cost $C_i$ is typically the driving factor in total cost. This becomes important when defining the actions for the warehouse agent in subsequent sections.}. As before, $Q_{\mathrm{waste},i}(nt)$ is the amount of product $i$ wasted within the warehouse in the $n$ time periods from $(nt-n)$ to $(nt-1)$.

\subsection{System goal} \label{subsec:goal}

The goal of the supply chain (or the portion that we are interested in) is to minimise the long term cost,
\begin{equation}
    G_t = \sum_{k=0}^{\infty} \gamma^k \left[ g_1\,\mathcal{C}_\mathrm{wh}(t+k) + g_2\,\mathcal{C}_\mathrm{st}(t+k) \right],
    \label{eq:optimise}
\end{equation}
where $g_1$ and $g_2$ are constant weights, $\gamma$ is a discount factor, $\mathcal{C}_\mathrm{st}=\sum_j \mathcal{C}_{j,\mathrm{st}}$, and we assume that $\mathcal{C}_\mathrm{wh}(t+k)=0$ for time periods that are not multiples of $n$. The first term in $G_t$ is the cost of purchasing products from vendors to replenish the warehouse, while the second term is the total cost incurred by stores in the current time period. Each store $j$ is to compute a local mapping $({\bf{x}}_j(t), \hat{{\bf{w}}}_j(t+1), \chi (t)) \rightarrow {\bf{u}}_j(t)$ that maximises $G_t$ without using information about the inventory levels or sales forecasts of other stores. The warehouse is to compute a mapping $({\bf{X}}(nt), \chi (nt), \hat{\chi}(n(t+1)), \hat{\bf{W}}(n(t+2))) \rightarrow \mu (nt)$, where ${\bf{X}}(nt)$ is a $P\times S$ matrix containing the inventory levels of all products in each store at time $nt$.


\section{Solution Methodology} \label{sec:method}

The description in Section \ref{subsec:goal} shows that the problem can be modeled as a Markov Decision Process ($\mathcal{S}, \mathcal{A}, \mathcal{T}, \mathcal{R}, \gamma$), where,
\begin{itemize}
    \item $\mathcal{S}$ represents the state space defined by some combination of the states ${\bf{x}}_j$, $\hat{{\bf{w}}}_j$, $\chi$, $\hat{\chi}$, ${\bf{X}}$, and  $\hat{\bf{W}}$,
    \item $\mathcal{A}$ denotes the action space ${\bf{u}}_j$ or $\mu$, 
    \item $\mathcal{T}$ represents transition probabilities from one combination of state and action to the next,
    \item $\mathcal{R}$ denotes the rewards based on the cost minimisation objective, and 
    \item $\gamma$ is the discount factor for future rewards.
\end{itemize}
In this section, we describe RL based approach developed for store and warehouse replenishment. 
We use model-free RL rather than model-based algorithms for both the store and the warehouse, in order to avoid having to estimate the transition probabilities $\mathcal{T}$. Given the interdependence of all the products on each other due to the system constraints, estimating transition probabilities is a difficult task. 


The supply chain network topology depicted in Fig. \ref{fig:sc-eg} contains a natural hierarchy, with a single warehouse supplying multiple stores with products. However, the problem is not in typical hierarchical reinforcement learning form [\cite{vezhnevets2017feudal,nachum2018data}], in the sense of the higher level agent (warehouse) setting sub-goals for the lower level agents (stores) for achieving system objectives. The warehouse decisions affect the cost term $\mathcal{C}_\mathrm{wh}$ in (\ref{eq:optimise}) directly, and the term $\mathcal{C}_\mathrm{st}$ indirectly. The latter effect is a result of warehouse replenishment setting a maximum limit on store inventory levels, in turn affecting the terms in (\ref{eq:invreward}). As a result, this can be multi-agent settings while sharing rewards among each other. Similar to \textit{team spirit} proposed in [\cite{berner2019dota}], Warehouse replenishment agent would receive some proportion of the store reward/cost. However we do not anneal it and keep it constant over the training period. 

\begin{figure}
    \centering
    \includegraphics[width=0.8\textwidth]{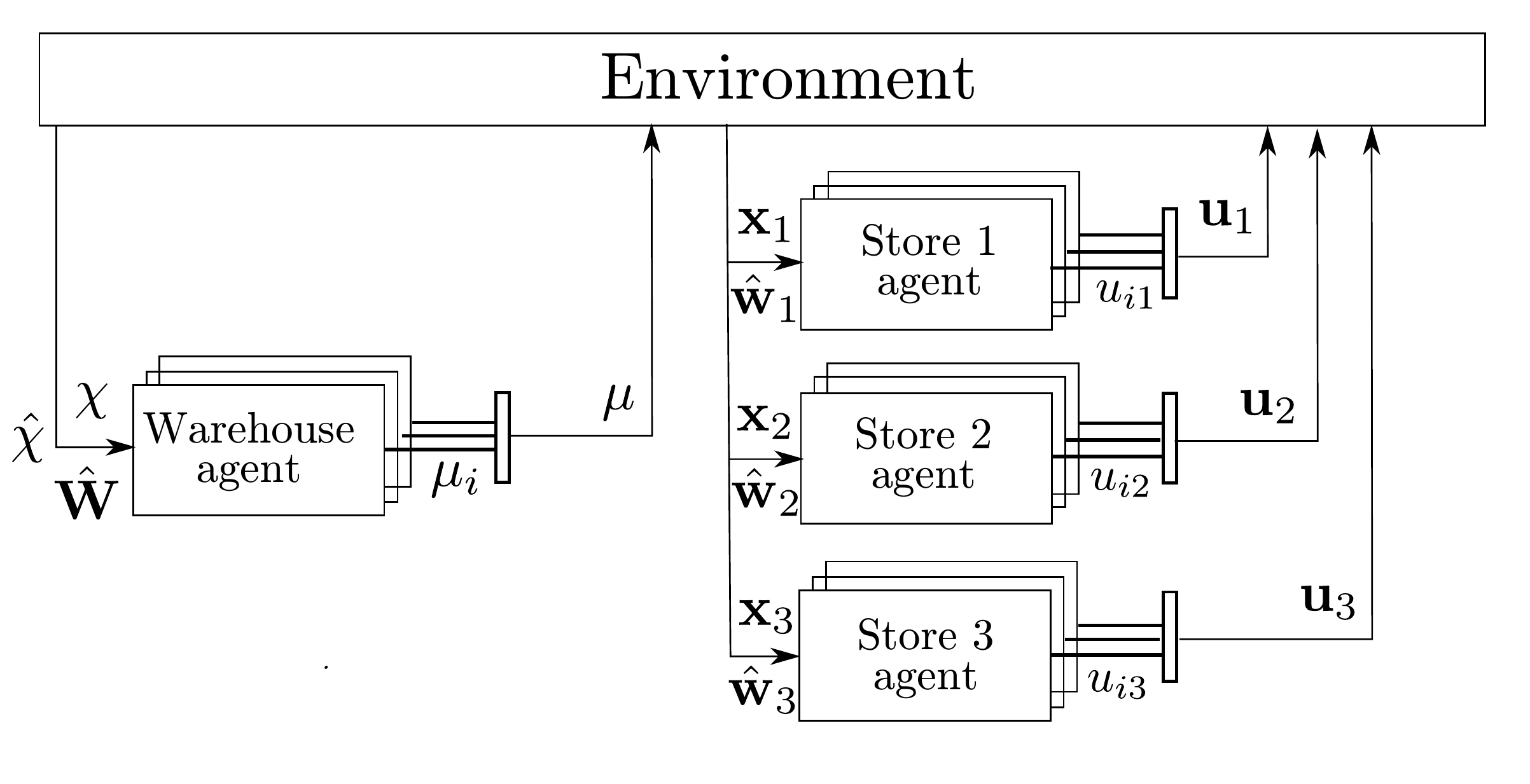}
    \caption{Multi-agent decision-making architecture. The store agents compute individual decisions $u_{ij}$, which are concatenated and clipped to conform to capacity constraints, before communicating them to the environment. A similar approach is used in the warehouse.}
    \label{fig:marl-arch}
\end{figure}

On the other hand, the store replenishment decisions affect $\mathcal{C}_\mathrm{st}$ directly, but are not strong drivers of $\mathcal{C}_\mathrm{wh}$. While higher demand from stores can drive up warehouse costs, there is a limit imposed by maximum inventory levels in the store. In the long run, average demand from stores to the warehouse should be approximately equal to the average sales from the stores, with the difference explained by wastage of perishables. Therefore, we design the inputs and rewards of the store agents to be independent of performance of the warehouse and of the other stores\footnote{We experimented with the alternative option of including warehouse state and rewards in the store agents, with surprising results. These are covered in Section \ref{sec:results}.}. The warehouse receives a portion of rewards from the stores in order to learn strategies that meet customer demand while minimising replenishment costs. The key design criterion in both types of agents is to ensure that the state and action spaces are independent of (i) the number $P$ of products moving through the supply chain, and (ii) the number $S$ of stores being supplied from the warehouse. The architecture depicted in Fig. \ref{fig:marl-arch} is designed to achieve this objective, and is described in detail below.

\subsection{Observations and actions}

Table \ref{tab:store_feat} summarises the observation and action spaces used for the store replenishment agent. The design of these spaces is inspired by the existing features that state-of-the-art replenishment systems use, and is meant to be `trustworthy' or `explainable' from a business perspective. The framework also allows us to compute replenishment quantities for each product independently, while receiving some global information in the form of aggregated demand on the system. 

\begin{table}[h]
\caption{Observations (inputs) and actions (outputs) for product $i$ in store $j$.}
\label{tab:store_feat}
\begin{center}
\begin{tabular}{|C{1.5cm}|l|}
\hline
Input & Explanation \\
\hline
$x_{ij}(t)$ & Current inventory level \\
$\hat{w}_{ij}(t+1)$ & Forecast sales in $[t,t+1)$ \\
$\hat{q}_{ij}(t)$ & Predicted wastage in existing inventory in $[t,t+1)$ \\
$v_i$ & Unit volume of product $i$ \\
${\bf{v}}^T\,{\hat{\bf{w}}}_j(t)$ & Total volume of forecast for all products \\
\hline \hline
Output & Explanation \\
\hline
$u_{ij}(t)$ & Replenishment quantity, quantised to a set of discrete choices \\
\hline
\end{tabular}
\end{center}
\end{table}

The first three inputs relate to the instantaneous state of the system with respect to product $i$ in store $j$. Of these, the current inventory level and sales forecast are simply the relevant elements of the vectors $\bf{x}$ and $\bf{w}$. The third input is an estimate of wastage of product $i$ in the next time period, based on its current inventory level. This number may be computed using probabilistic estimates (in the case of fresh produce) or actual sell-by dates (in the case of labelled products). It tells the algorithm about the expected loss of inventory in addition to the sales forecast. The fourth input is unit volume of product $i$, which is the relevant aspect of product meta-data in this model\footnote{Where relevant, one could also consider unit weight and physical dimensions.}. It helps the algorithm recognise the effect of the current product on the aggregate capacity constraint. 

The last input in Table \ref{tab:store_feat} is a derived feature that indicates total demand on the system, with respect to constraint (\ref{eq:truckvol}). This indicator acts as an inhibitor to the replenishment quantity for product $i$, if the total demand on the system is high. It also helps the RL agent correlate the last term in the observed rewards (the capacity exceedance penalty) with the observations. The output of the RL agent is ${u}_{ij}(t)$, which is the desired action for product $i$ in store $j$ at time $t$. Individual actions are concatenated to form ${\bf{u}}_j(t)$, as shown in Fig. \ref{fig:marl-arch}. In addition to the concatenation, we also normalise the vector by $\rho_j$ in order to ensure compliance with constraint (\ref{eq:truckvol}). For reasons explained in Section \ref{sec:results}, we do not include any inputs indicating the states of the warehouse or of other stores.

Table \ref{tab:dc_feat} summarises the observation and action spaces used for the warehouse replenishment agent. The first three inputs relate to the instantaneous state of the system with respect to each product where $\hat{\bold{W}}_{i}(n(t+2))$ is the summation of the each store order normalized to DC maximum shelf capacity and $\hat{\chi_{i}}(n(t+1))$ is the projected warehouse inventory at the end of $n(t+1)$ time period computed by subtracting the store order forecast $\hat{\bold{W}}_{i}(n(t+1))$ from the current inventory $\chi_{i}(nt)$. The fourth input is the predicted wastage quantity $\hat{Q}_{\mathrm{waste},i}(n(t+2))$ of product $i$ in the next time period, based on its projected inventory level. The last term $I_{\mathrm{empty}}$ is a boolean indicator stating whether a product is at empty stock. Given that (i) the fixed cost of warehouse replenishment (ordering from a vendor) is the driving factor behind total cost, and (ii) there is no physical constraint on total replenishment of the warehouse (unlike in the case of stores), we define the action of the warehouse agent to be the binary indicator $b_i$ from Equation (\ref{eq:ware_cost}). The quantity of replenishment $\mu_i(nt)$ is computed to be equal to the remaining space for that product,
\begin{equation*}
    \mu_i(nt) = \begin{cases} 1 - \chi_i(nt), & \text{ if }b_i(nt) = 1 \\
                              0, & \text{ if }b_i(nt) = 0 \end{cases}
\end{equation*}

\begin{table}[h]
	\caption{Observations (inputs) and actions (outputs) for product $i$ in warehouse.}
	\label{tab:dc_feat}
	\begin{center}
		\begin{tabular}{|C{3cm}|l|}
			\hline
			Input & Explanation \\
			\hline
			$\chi_{i}(nt)$ & Current inventory level \\
			$\hat{\chi_{i}}(n(t+1))$ & Projected inventory level when action will be implemented \\
			$\hat{\bold{W}}_{i}(n(t+2))$ & Forecast aggregate store replenishment demand \\
			$\hat{Q}_{\mathrm{waste},i}(n(t+2))$ & Predicted wastage quantity \\
			$I_{\mathrm{empty}}$ & Binary flag indicating whether inventory is at empty stock\\
			\hline \hline
			Output & Explanation \\
			\hline
			$b_{i}(nt)$ & Replenishment decision, binary \\
			\hline
		\end{tabular}
	\end{center}
\end{table}

\subsection{Rewards}

The key challenges with computation of individual elements of ${\bf{u}}_j$ are (i) ensuring that the system-level constraint (\ref{eq:truckvol}) is met, and (ii) that all products are treated fairly. Both challenges are partially addressed using the reward structure. The fairness issue is addressed using the percentile spread term in (\ref{eq:invreward}), since it penalises the agent if some products have low inventories while others are at high levels. The volume constraint is introduced as a soft penalty in the following reward definition, adapted for individual decision-making.
\begin{equation}
R_{ij}(t) = 1 - b_{ij,\mathrm{empty}}(t) - q_{\mathrm{waste},ij}(t) - \Delta {\bf{x}}_j(t)^\mathrm{.95}_{\mathrm{.05}} - \alpha\, (\rho_j - 1), \label{eq:itemreward}
\end{equation}
where $b_{ij,\mathrm{empty}}(t)$ is a binary variable indicating whether inventory of product $i$ in store $j$ dropped to zero in the current time period, $\alpha$ is a constant parameter, and $\rho_j$ is the ratio of total volume requested by the RL agent to the volume that can be feasibly transported to store $j$. We formally define this as,
\begin{equation*}
    \rho_j = \max \left( \frac{{\bf{v}}^T\,{\bf{u}}_j(t)}{v_{\mathrm{max},j}},\; 1 \right).
\end{equation*}
Equation (\ref{eq:itemreward}) defines the reward that is actually returned to the RL agent, and is related to the store cost defined in (\ref{eq:invreward}). If the aggregate actions output by the agent (across all products) do not exceed the available capacity ($\rho = 1$), then the average value of (\ref{eq:itemreward}) is equal to $(1-\mathcal{C}_{j,\mathrm{st}})$. This implies that maximising $R_{ij}(t)$ is equivalent to minimising $\mathcal{C}_{j,\mathrm{st}}$, as long as system constraints are not violated. The last two terms of (\ref{eq:itemreward}) are common to all products at a given time step $t$.

The warehouse reward is driven primarily by the cost of replenishment $\mathcal{C}_{\mathrm{wh},i}$ and the quantity of replenishment requests from stores that are refused in the time periods $n(t+1):n(t+2)$, denoted by $\Omega_i(n(t+1))$. Note that the refused order penalty applies only from time period $n(t+1)$ onwards, because this is when the effect of action $b_i(nt)$ is seen. In addition to the warehouse cost, we also include a proportion of the store replenishment reward to the warehouse agent. The net reward function is given by,
\begin{equation}
R_{\mathrm{wh},i}(nt) = 1 - \mathcal{C}_{\mathrm{wh},i}(nt) - \alpha_1 \, \Omega_{i}(n(t+1)) + \alpha_2 \sum_{j=1}^{S}R_{ij}(t),
\label{eq:dc_reward}
\end{equation}
%
%
where $\alpha_1$ and $\alpha_2$ are user-defined constant weights.





\subsection{Model implementations for store and warehouse agents}

We now describe the specific RL algorithms and neural network architectures for the store and warehouse agents. While both agents use A2C [\cite{konda2000actor}], the store agent uses a modified training procedure for faster convergence and smoother policies.




\subsubsection{Store Agent} \label{subsec:storearch}

The computation of each element $u_{ij}(t)$ is carried out using advantage actor critic (A2C) [\cite{konda2000actor}]. The critic network accepts the 5 features from Table \ref{tab:store_feat} as input, contains one hidden layer with \textit{relu} as activation, and produces a scalar value output with \textit{linear} activation in the output layer. Actor model has a input same as critic with 3 hidden layers with \textit{relu} as activation and final layer with \textit{Softmax} user-defined set of 14 quantized actions between 0 (no replenishment) and 1 (replenishment equal to maximum space available for product $i$ in store $j$)\footnote{The specific actions used in this paper are $\{ 0, 0.005, 0.01, 0.0125, 0.015, 0.0175, 0.02, 0.03,$ $0.04, 0.08, 0.12, 0.2, 0.5, 1 \}$}. In general, the number of quantized actions is arbitrary, and can be based on the most common replenishment quanta in the store\footnote{For example, bottles of water may be easiest to ship in multiples of 12.}.

Weights are shared across the products to ensure the fairness across products and to ensure that new products could be added without further fine-tuning of the weights. The advantage of this approach is that it splits the original task into constant-scale sub-tasks that can be executed in parallel.  Therefore, the same algorithm can be applied to instances where there are a very large (or variable) number of products. However, we create separate sets of models for each store to better model different behaviours of that store. This is a realistic situation in dynamic environments; for example, retailers are continuously adding new products and removing old products from their portfolio.

\begin{figure}[t]
    \centering
    \includegraphics[width=0.8\textwidth]{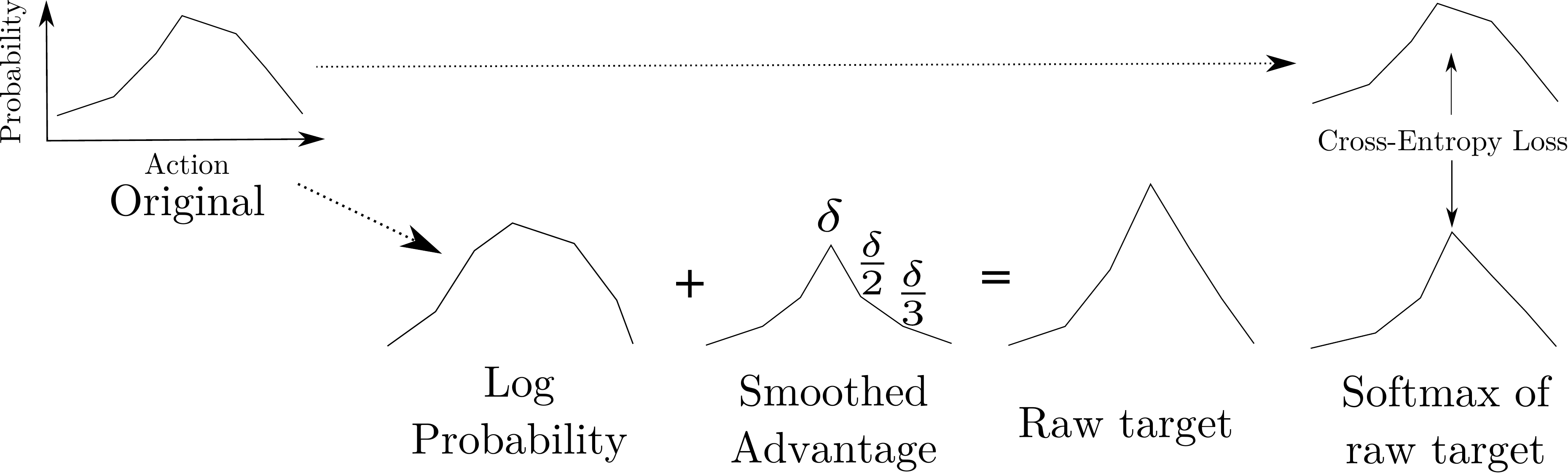}
    \caption{Training procedure for the actor network in the store replenishment agent.}
    \label{fig:AC1}
\end{figure}

Training paradigm and loss functions of this agent is modified from vanilla A2C~[\cite{konda2000actor}], For each action $u_{ij}$, we first compute the advantage $\delta_{ij}$ using the TD(0) discounted reward, as per standard A2C. However, we do not directly use the advantage-weighted sum of logits to compute the gradients of the actor, because of the following reasoning. First, our output layer is \textit{softmax}. Additionally, unlike the typical discrete choice problem, the actions in our case represent quantised levels of replenishment. Therefore, consecutive actions have a similar effect on the trajectory, and the distinction between \textit{right} and \textit{wrong} actions is not as stark as in discrete choice. We account for this property by using the following procedure for training the actor, which we found benefits both the rate of learning and the smoothness of learned policies.
\begin{enumerate}
\item Let us denote by $\ell_{ij}$ the vector of logits corresponding to the softmax output for product $i$ in store $j$. Further, let us assume that the $r^\mathrm{th}$ action is chosen, and it has the corresponding logit activation $\ell_{ij}(r)$.
\item We denote by $\ell^*$ the target logit vector that we wish to match. The elements of $\ell^*$ are computed using the relation,
\begin{equation*}
\ell^*_{ij}(r') = \left(\ell_{ij}(r')+\frac{\delta_{ij}}{|r-r'|+1}\right)\;\;\forall r'\in \{1,2,\ldots,14\}.
\end{equation*}
\item The above relationship increases the $r^\mathrm{th}$ logit by $\delta_{ij}$, but also adds fractions of the advantage to neighbouring actions. This results in a smoother target vector. Finally, the target vector $\ell^*$ is run through a \textit{softmax} output. The actor network is trained to minimise the cross-entropy loss between the original and the modified \textit{softmax} output. This is shown graphically in Fig. \ref{fig:AC1}.
\end{enumerate}
%

\subsubsection{Warehouse Agent} \label{subsec:warehousearch}


We describe an algorithm for the  warehouse replenishment agent which learns to handle multiple stores with multiple capacity and manage to learn the binary actions $\mu_{i}$ of whether to replenish or not at that point of time. The lead time for DC is 1 day or after n time period $(nt)$ of store replenishment, hence the actions need to be taken one day advance of its actual replenishment, which includes stochasticity while predicting the projected inventory value rather than actual inventory at that time. The agent parameters is shared across all the products of the same warehouse similar to store agents. DC/Warehouse follows vanilla A2C algorithm as its output is only binary action. The actor consists of two layer network with \textit{relu} as hidden activations and \textit{Softmax} as the output activation. Critic has single hidden layer with \textit{relu} activation for hidden layer while the output layer activation is \textit{linear}. Both actor and critic network accepts the input features as given in Table \ref{tab:dc_feat}. 

\subsection{Training process}

Considering the store and warehouse replenishment as a two-echelon supply chain problem, we train the store agents (lower echelon) first, followed by the warehouse agent (higher echelon). The specific steps that we follow are listed below, with the specific example of one warehouse supplying three stores, and with each store being replenished by a separate truck (volume constraint).
\begin{enumerate}
    \item Training of store replenishment agents
        \begin{itemize}
            \item Three copies of the A2C architecture for store replenishment (Section \ref{subsec:storearch}) are created, one for each store. Weights for all products within a given store are shared. For simplicity, we assume that all three stores handle the same products (50, 220, or 1000, depending on the instance). However, this is not a technical requirement.
            \item The three stores are trained independently of each other (on separate computational threads), using transactional data derived for each store (described in Section \ref{subsec:data}). 
            \item During store training, we assume that requested replenishment quantities for all products are always available in the warehouse (see note below). Training is stopped upon reaching convergence, as explained in Section \ref{sec:results}.
        \end{itemize}
    \item Training of warehouse replenishment agent
        \begin{itemize}
            \item Warehouse agent training starts only after all three stores have converged. During this phase, we do not train the store agents further. The goal is to compute a warehouse replenishment policy that is able to serve the store demand (which is the primary objective of the supply chain), while minimising its own cost subject to that constraint.
            \item We initialise the warehouse A2C architecture (\ref{subsec:warehousearch}), and load the previously trained store replenishment models. Since the warehouse replenishment is dependent on store sales, we reuse the product sales data used for store agent training. 
            \item Since the warehouse inputs include $\hat{\chi_{i}}(n(t+1))$, the projected warehouse inventory at the time action $\mu(nt)$ takes effect, and $\hat{\bold{W}}_{i}(n(t+2))$ forecast of replenishment requests from stores in the subsequent $n$ time periods, we need a way of computing these quantities. Therefore, we run two environments in parallel: one for the simulation and training, and another for the rollout of store replenishment. In each time step, the rollout environment is initialised with the current inventory levels in the stores and the warehouse, and the forecast data are used to compute $\hat{\chi_{i}}(n(t+1))$ and $\hat{\bold{W}}_{i}(n(t+2))$.
        \end{itemize}
\end{enumerate}

\begin{figure}[h]
  \centering
  \includegraphics[width=0.9\textwidth]{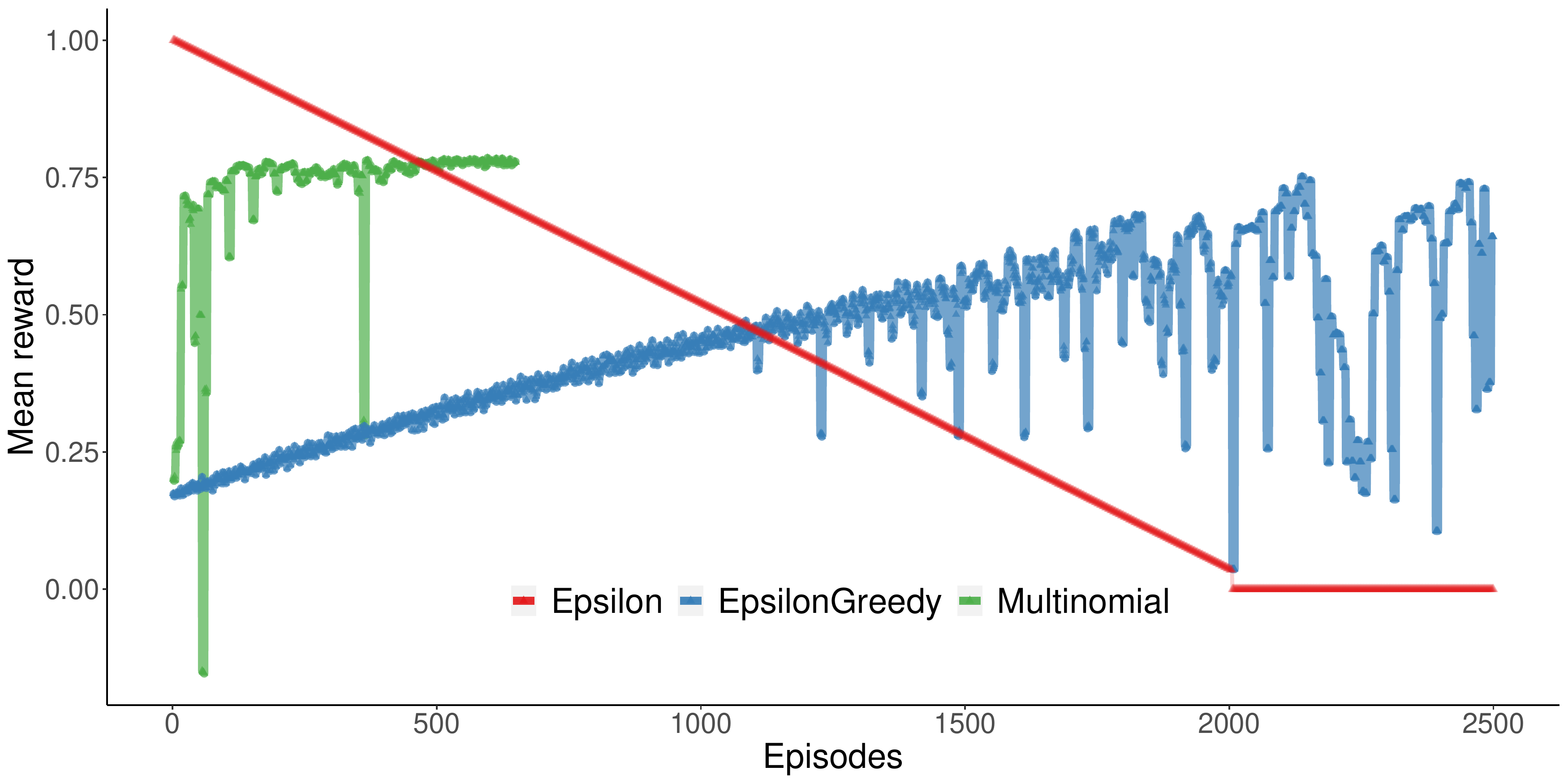}
  \caption{Comparison of epsilon greedy training with the multinomial approach}
  \label{fig:epsilon_greedy}
\end{figure}

%
The output actions for store agent as well as warehouse agent are drawn from a multinomial distribution. The multinomial distribution is a multivariate generalisation of the binomial distribution. Since it randomly draws samples it includes the exploration required as in case of epsilon greedy method until convergence. We have observed that multinomial exploration tends to converge faster as compared to the epsilon greedy approach as shown in Figure \ref{fig:epsilon_greedy}. Epsilon linearly decays over time in 2000 episodes where we exploit the current situation with probability $(1 - epsilon)$ and explore a new option randomly with probability $epsilon$, where epsilon is a hyper-parameter. As noticeable, the learning does not converge in epsilon greedy approach even after 2500 episodes while using multinomial the learning curve converges within 500 episodes.

\section{Experiments and Results} \label{sec:results}

We describe experiments with a scenario consisting of primarily one warehouse and three stores. Three different instances are considered, with 50, 220, and 1000 product types respectively. Towards the end of this section, we also describe experiments where the algorithms are trained on a certain number of stores and products, and tested on a different number of stores or products.

\subsection{Data Description} \label{subsec:data}

We use a public data set for brick and mortar stores [\cite{Kaggle}] as the basis for the experimentation. The original data set includes purchase data for 50,000 product types and 60,000 unique customers. However, it does not contain meta-data about the products (dimensions, weight) and also does not specify date of purchase (although it specifies the day of week). Instead, it measures the number of days elapsed between successive purchases by each customer. We first assign a random date to the first order of each unique customer while respecting the day of week given in the original data set. This implicitly assigns specific date and time stamps to each purchase. Additionally, we assign dimension and weight to each product type based on the product label (which is available in the original data set). Hence the modified dataset consists of forecasts and corresponding sales for each product for 349 days, equivalent to 1396 time periods of 6 hours each. 
We divide the dataset into training and testing sets, with the initial 900 time periods as training set and the remaining 496 time periods as the testing set.


We evaluated our approach on three subsets of the prepared dataset, consisting of 50, 220 and 1000 products respectively. The products are assumed to pass through a single warehouse, which supplies three stores using separate trucks (separate volume constraints). While all the stores handle the same variety of products, they each have different shelf capacities, and different sales rates relative to those capacities. The sales statistics and relative sizes of the stores are listed in Table \ref{tab:trcuk_vol_table}. For each instance, the warehouse capacity was sized to hold about 7 days (28 time periods) worth of demand from the stores, aggregated over all three stores.

\begin{table}[b]
\caption{Summary of the three data sets, listing for each store: (i) average sales per time period normalized by shelf capacities of the products, (ii) relative shelf capacities of the stores, and (iii) chosen truck volume.}
\centering
  \label{tab:trcuk_vol_table}
  \begin{tabular}{|c|c|c|c|c|c|c|c|c|c|} \hline
    \multirow{3}{*}{ID} & \multicolumn{3}{c|}{50 products} & \multicolumn{3}{c|}{220 products} & \multicolumn{3}{c|}{1000 products} \\
    
    & Norm & Rel & Trk vol & Norm & Rel & Trk vol & Norm & Rel & Trk vol \\
    & Sales & Capc & $10^6$cm$^3$ & Sales & Capc & $10^6$cm$^3$ & Sales & Capc & $10^6$cm$^3$ \\
    \hline
    Store1 & 1.7\% & 1x & 0.45 & 2.1\% & 1x & 1.80 & 1.9\% & 1x & 2.20 \\
    Store2 & 3.3\% & 2x & 1.30 & 3.7\% & 2x & 6.20 & 2.7\% & 1.5x & 4.20 \\
    Store3 & 3.3\% & 1.5x & 1.00 & 3.7\% & 1.5x & 4.70 & 2.8\% & 2x & 5.50 \\  \hline 
\end{tabular}
\end{table}

\begin{figure}
    \centering
    \subfigure[]{\includegraphics[width=0.33\textwidth]{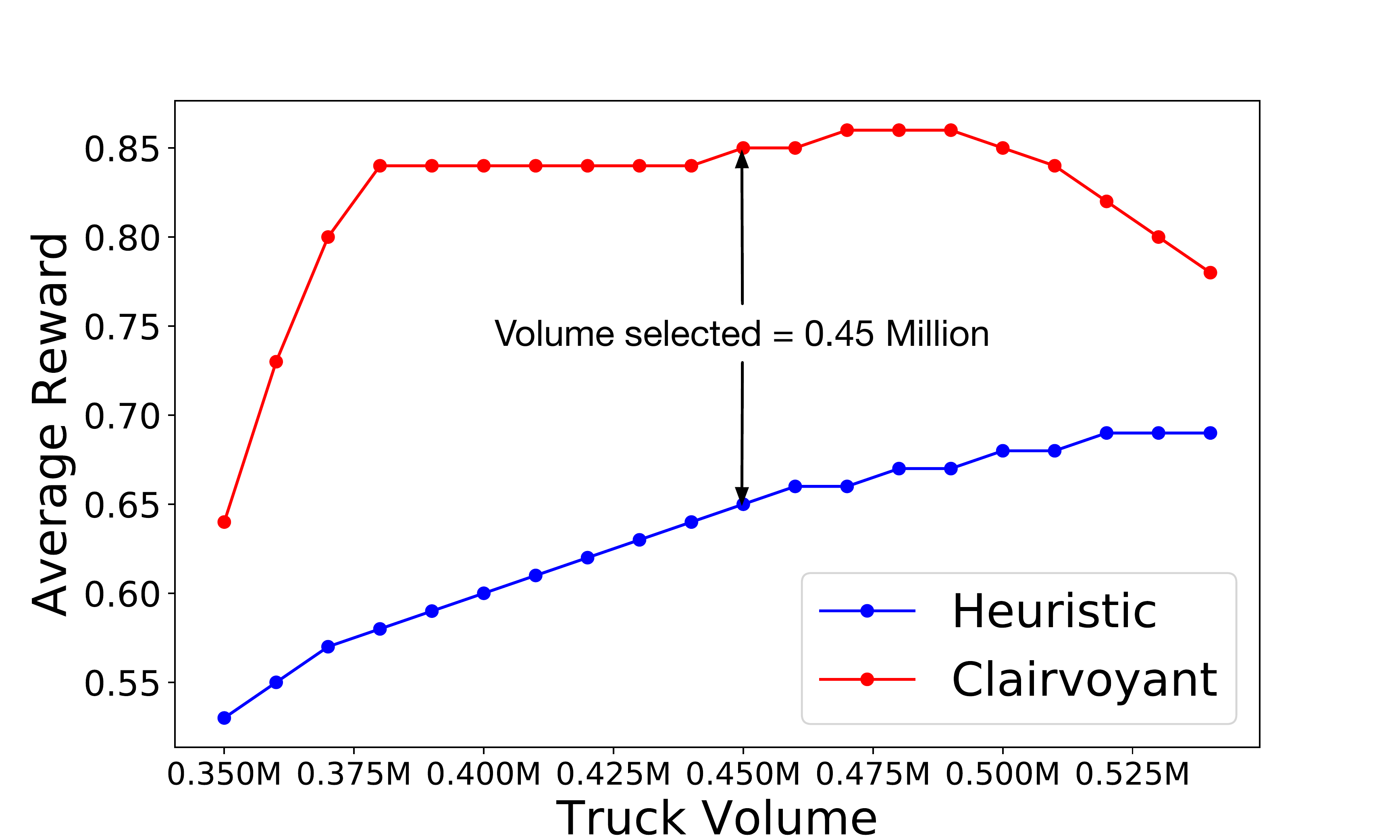}} 
    \subfigure[]{\includegraphics[width=0.33\textwidth]{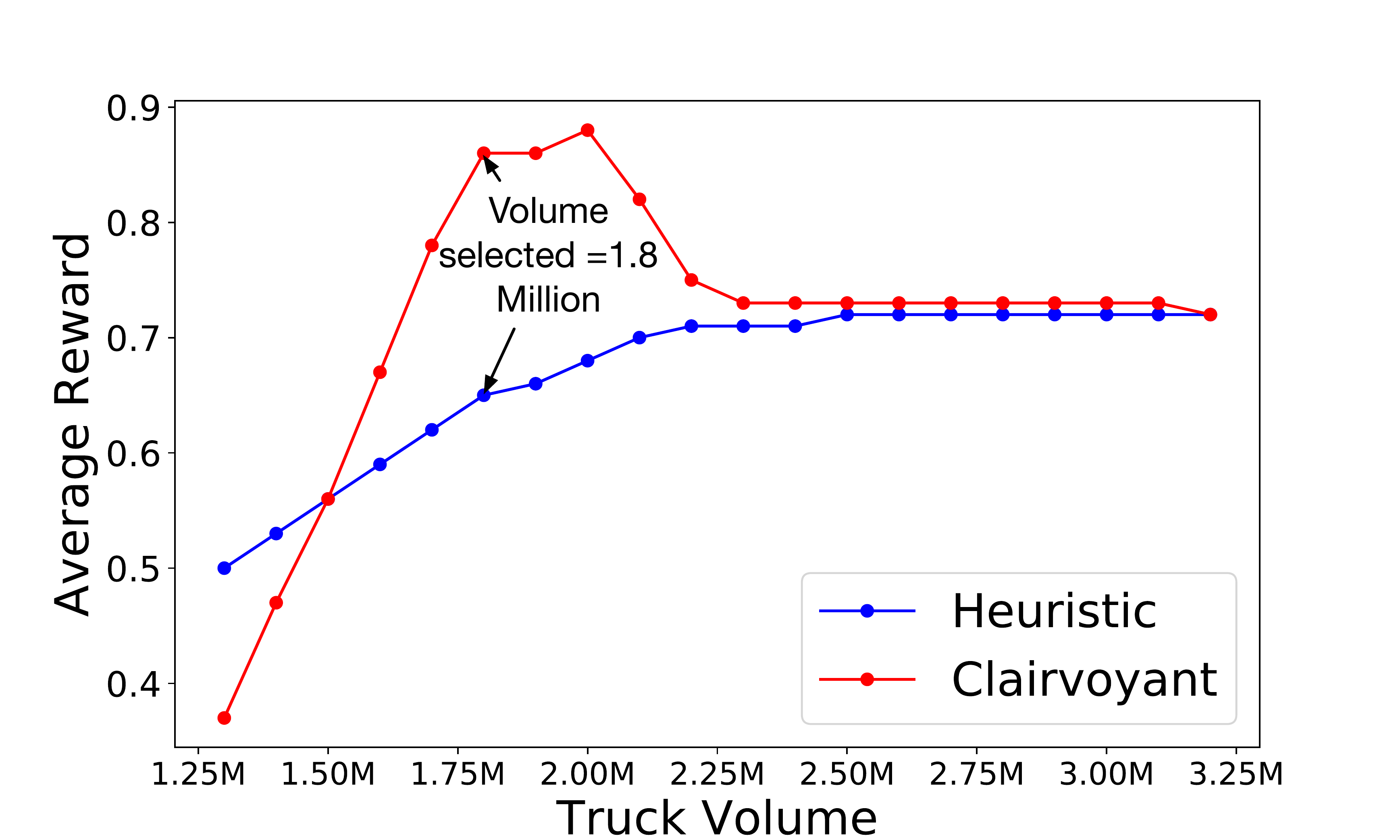}} 
    \subfigure[]{\includegraphics[width=0.32\textwidth]{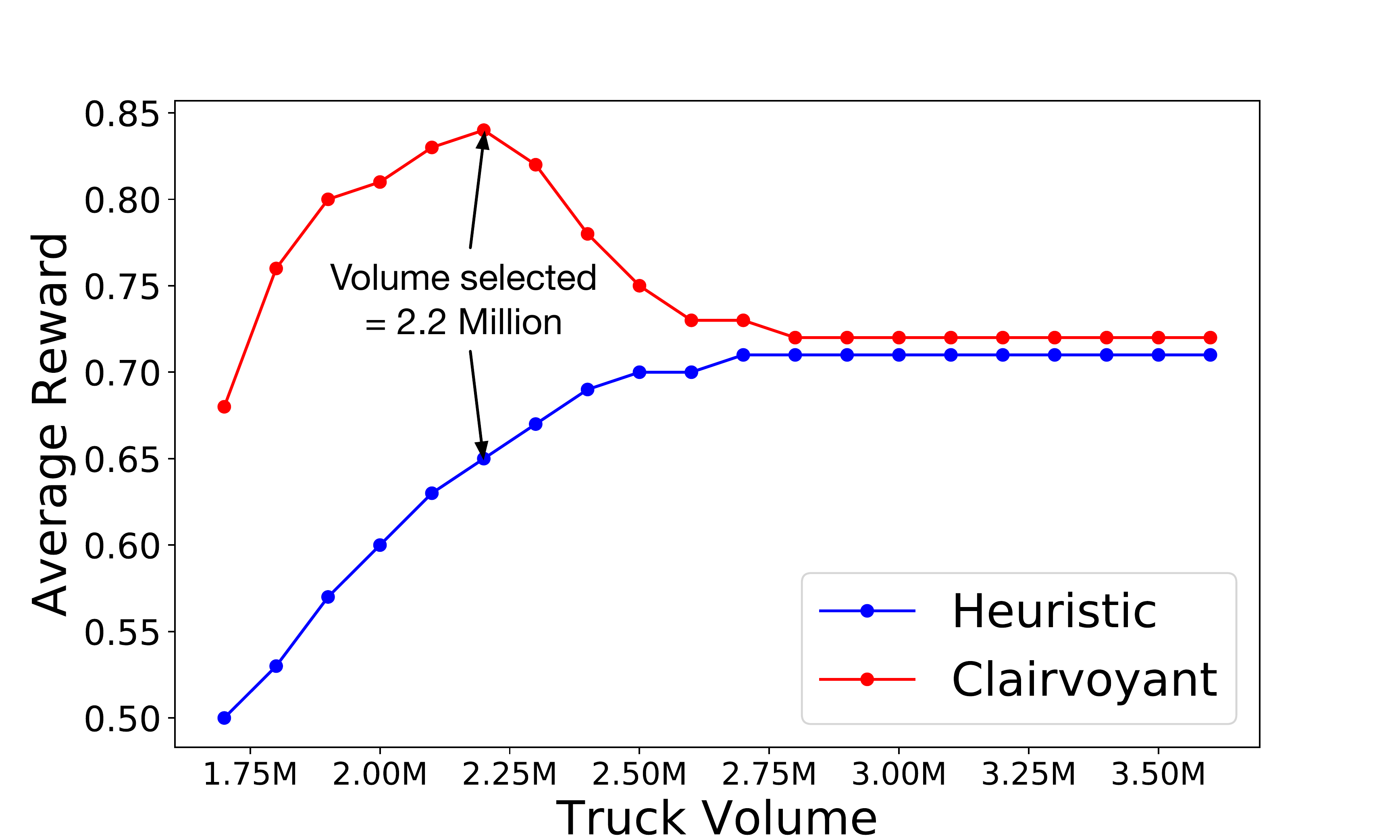}}
    \caption{Truck volume and weight selection by validating across a range of volume and weight(a) Store 1 (50 products) (b) Store 1 (220 products) (c) Store 1 (1000 products)}
    \label{fig:truck_volume}
\end{figure}

In order to set the volume capacity of each truck in each instance, we wished to ensure a value that was high enough for the stores to maintain modest inventory levels of all products, but low enough to ensure that mistakes in replenishment resulted in poor performance for several subsequent time periods. We ran two heuristic baseline algorithms (described in detail later) for different values of the volume constraint. Illustrative obtained rewards are shown in Fig. \ref{fig:truck_volume}. The `clairvoyant' version uses actual sales data (perfect forecasts) for replenishment decisions, and thus significantly outperforms the basic heuristic over a range of truck volume constraints. We chose values where sufficient scope for optimisation was available, between the performance of the basic heuristic and its clairvoyant variant.


\begin{figure}
    \centering
    \subfigure[]{\includegraphics[width=0.32\textwidth]{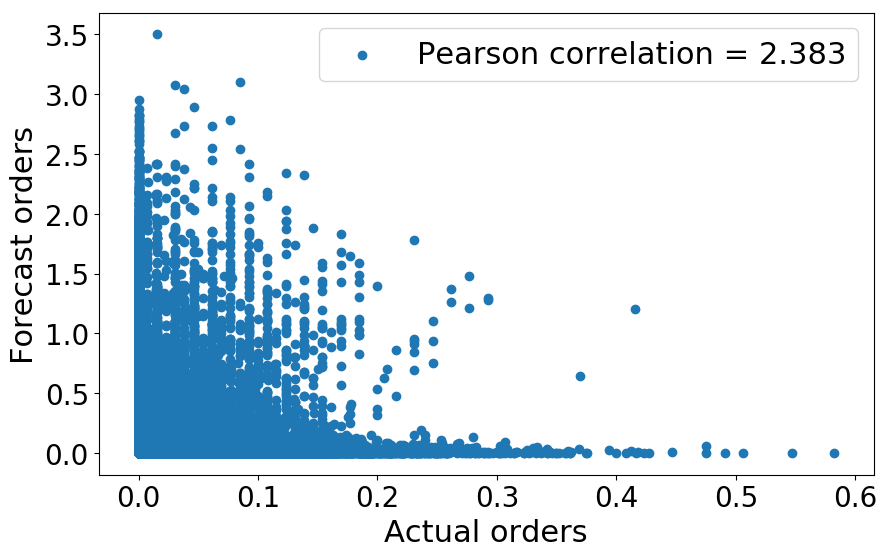}} 
    \subfigure[]{\includegraphics[width=0.32\textwidth]{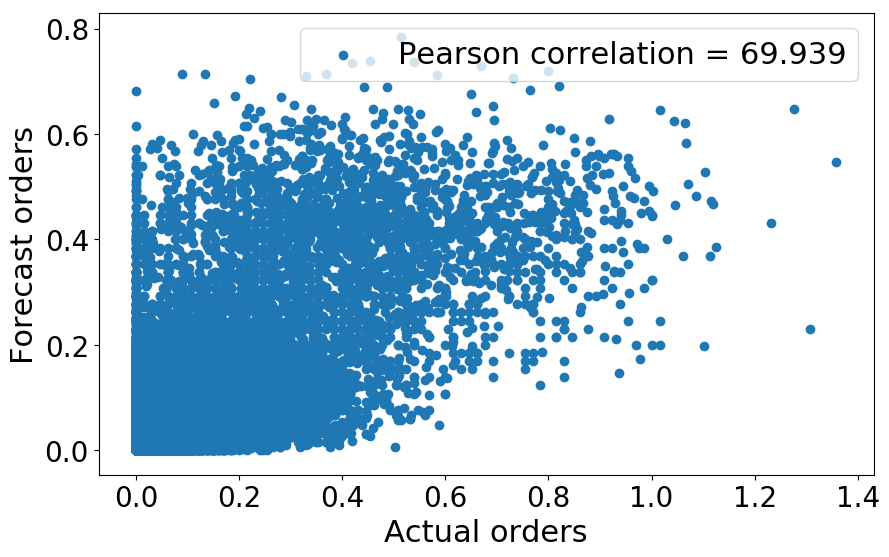}} 
    \subfigure[]{\includegraphics[width=0.32\textwidth]{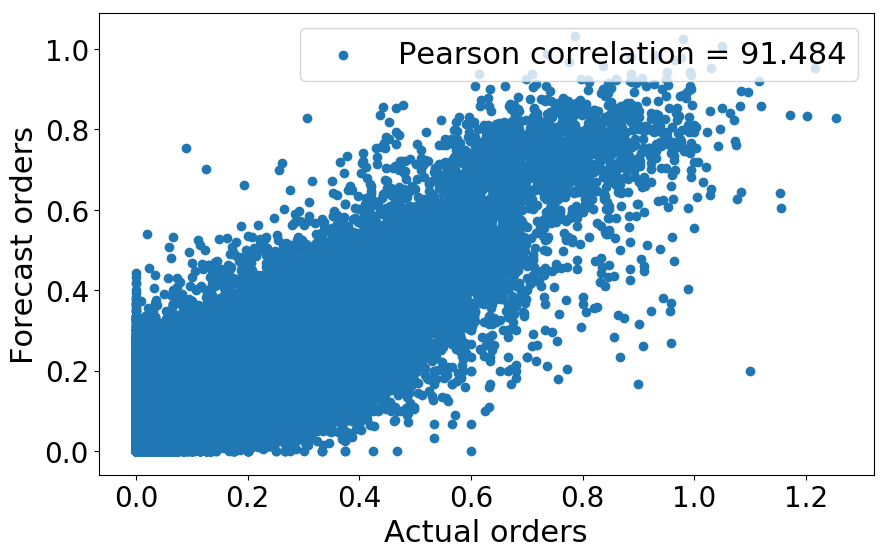}}
    \caption{Relationship between actual and forecast orders for (a) Store 1 (50 products) (b) Store 1 (220 products) (c) Store 1 (1000 products)}
    \label{fig:forecast_actual}
\end{figure}

Figure \ref{fig:forecast_actual} shows the relation between the forecast and the actual orders across the number of products for the three instances. The Pearson correlation measures the linear relationship between the forecast provided to the RL agent and the actual sales. The instance with 50 products is given poor forecast accuracy (nearly independent of actual sales), the instance with 220 products has reasonable accuracy, while the instance with 1000 products has very high forecast accuracy. The subsequent experimental results show that despite the difference in forecast accuracy across the various product datasets, the RL agent is able to learn good replenishment policies. 

\subsection{Baseline algorithms}

We use two versions of a standard algorithm from supply chain literature known as s-policy [\cite{nahmias1994optimizing}] for computing baseline levels of solution quality. We do not run other reinforcement learning baselines because our experiments showed that algorithms such as DQN [\cite{mnih2015human}] and DDPG [\cite{lillicrap2015continuous}] reach similar levels of performance, if the state and actions spaces are kept the same as those described in Section \ref{sec:method}. The only differences are minor variations in performance, and the time required for convergence. DDPG in particular, was noted to take longer to converge than either DQN or A2C. In the description below, both store and warehouse replenishment is carried out by similar heuristics.

\subsubsection{Constant-inventory heuristic}

Algorithms in the s-policy family of heuristics aim to maintain a constant level of inventory over time, by accounting for current inventory levels and the upcoming forecast demand. If we define $\bf{x}_j^*$ to be the \textit{target} level of inventories for the products in store $j$, the desired replenishment quantity is designed to satisfy forecast sales and reach $\bf{x}^*$ by the end of the time period. The expression is given by,
\begin{equation}
{\bf{u}}_{\mathrm{he},j}(t) = \max\left[ 0, {\bf{x}}^*_j + \hat{\bf{w}}_j(t+1) - {\bf{x}}_j(t)^-\right]. \label{eq:he_desired}
\end{equation}
The computed action ${\bf{u}}_\mathrm{he}(t)$ according to (\ref{eq:he_desired}) already satisfies constraints (\ref{eq:control}) and (\ref{eq:shelf}), since ${\bf{0}}\leq {\bf{x}}_j^* \leq {\bf{1}}$. As with the RL outputs, the desired replenishment quantities will be normalised by the total demand on volume capacity. The normalisation constant $\rho_{\mathrm{he},j}$ will be defined by,
\begin{equation*}
    \rho_{\mathrm{he},j} = \max \left( \frac{{\bf{v}}^T\,{\bf{u}}_\mathrm{he}(t)}{v_{\mathrm{max},j}},\; 1 \right)
\end{equation*}
In this work, we use a constant value of ${\bf{x}}^*_j = 0.25$ for all products and stores. For warehouse replenishment, we use the same concept, with a small modification to account for the binary nature of decisions $b_i$. We choose to replenish a product $i$ (that is, $b_i(nt)=1$) if the projected inventory level at time $n(t+1)$ is expected to be insufficient for lasting until time $n(t+2)$; that is, if $\hat{\chi}_i(n(t+1)) < \hat{\bf{W}}_i(n(t+2))$.

\subsubsection{Clairvoyant version of heuristic}

For a baseline with stronger solution quality, we define a `clairvoyant' version of the heuristic approach described above. This algorithm uses the same procedure as described for the constant-inventory heuristic for stores, but the forecast of product sales is replaced by the true realised future demand. Warehouse replenishment decisions are set to $b_i=1$ if the true replenishment demand from stores exceeds the true inventory at time $n(t+1)$. This is clearly infeasible from a practical perspective (sales cannot be predicted exactly), but it allows us to remove the effect of forecast inaccuracy from the results. As we show in the next section, clairvoyance gives the heuristic an advantage over RL which does not use future information, but not every time.

\subsection{Results and Discussion}

\begin{figure}
  \centering
  \includegraphics[width=0.9\textwidth]{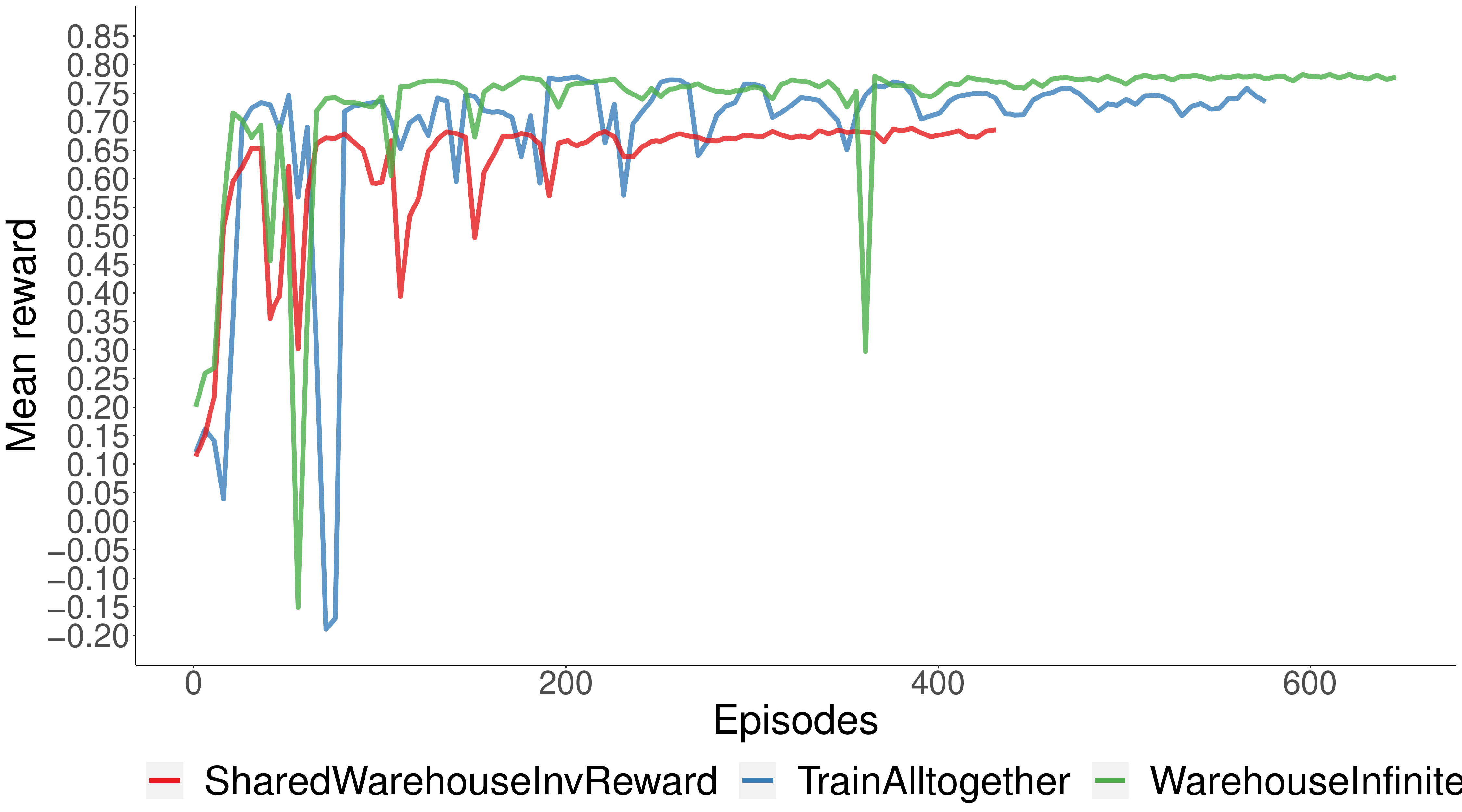}
  \caption{Rewards during training for Store 1 (220 products), during three different curricula: (i) independently of the warehouse (green plot), (ii) simultaneously with the warehouse and other stores (blue plot), and (iii) simultaneously with the warehouse and other stores, and including a portion of warehouse reward (red plot). Training is automatically terminated when convergence is achieved, as described in the text.}
  \label{fig:train_regime_comp}
\end{figure}

We now report and analyse results of training and testing, run for several instances of the problem. Wherever applicable, training of store agents is completed first, assuming the warehouse has infinite capacity. Subsequently, the warehouse agent is trained in conjunction with the store agents, where the policies of the latter are kept fixed. We experimented with several variations of the training regime, including ones where the warehouse inventory and reward was partially fed to the store agents, and the stores and warehouse were trained together. Fig. \ref{fig:train_regime_comp} shows the rewards during training for Store 1 (220 products), during three different curricula: (i) independently of the warehouse, (ii) simultaneously with the warehouse and other stores, and (iii) simultaneously with the warehouse and other stores, and including a portion of warehouse reward. We see that the best store reward is achieved in the first case, when the store trains without any warehouse constraints or rewards. The warehouse rewards show an opposite trend to the one for stores, but the learnt policies are undesirable, as explained below.

In general, we found that inclusion of warehouse rewards during store training resulted in policies where the warehouse purchased fewer products from vendors, and stores (cognizant of low inventory in the warehouse) asked for lower replenishment from the warehouse, thus achieving lower sales overall. Simultaneous training of the warehouse and stores, while not providing an explicit reward, nevertheless constrained the environment to supply only as much replenishment as the warehouse currently held. We finally decided that the best option was to divide the responsibility of the agents clearly; the stores must fulfil customer demand for products (assuming that the warehouse will be able to provide as much replenishment as needed), while the warehouse must ensure that replenishment requests from stores are met to the extent possible. Any cost optimisation done by the warehouse should not result in lower store replenishment.

\begin{figure}[h]
  \centering
  \includegraphics[width=0.9\textwidth]{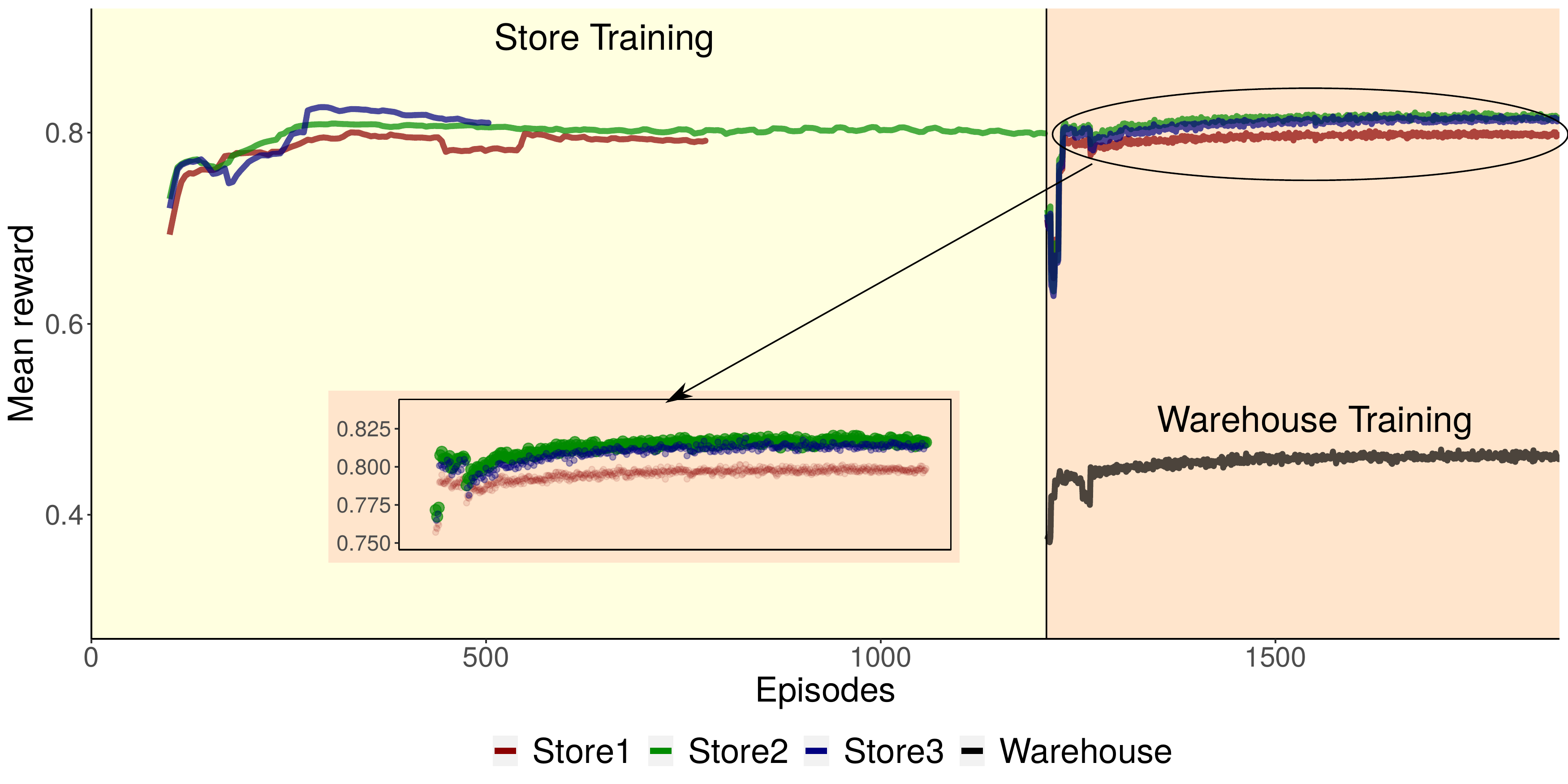}
  \caption{Training reward of Warehouse and Store for the 50 products}
  \label{fig:train_reward_50}
\end{figure}

\begin{figure}[h]
  \centering
  \includegraphics[width=0.9\textwidth]{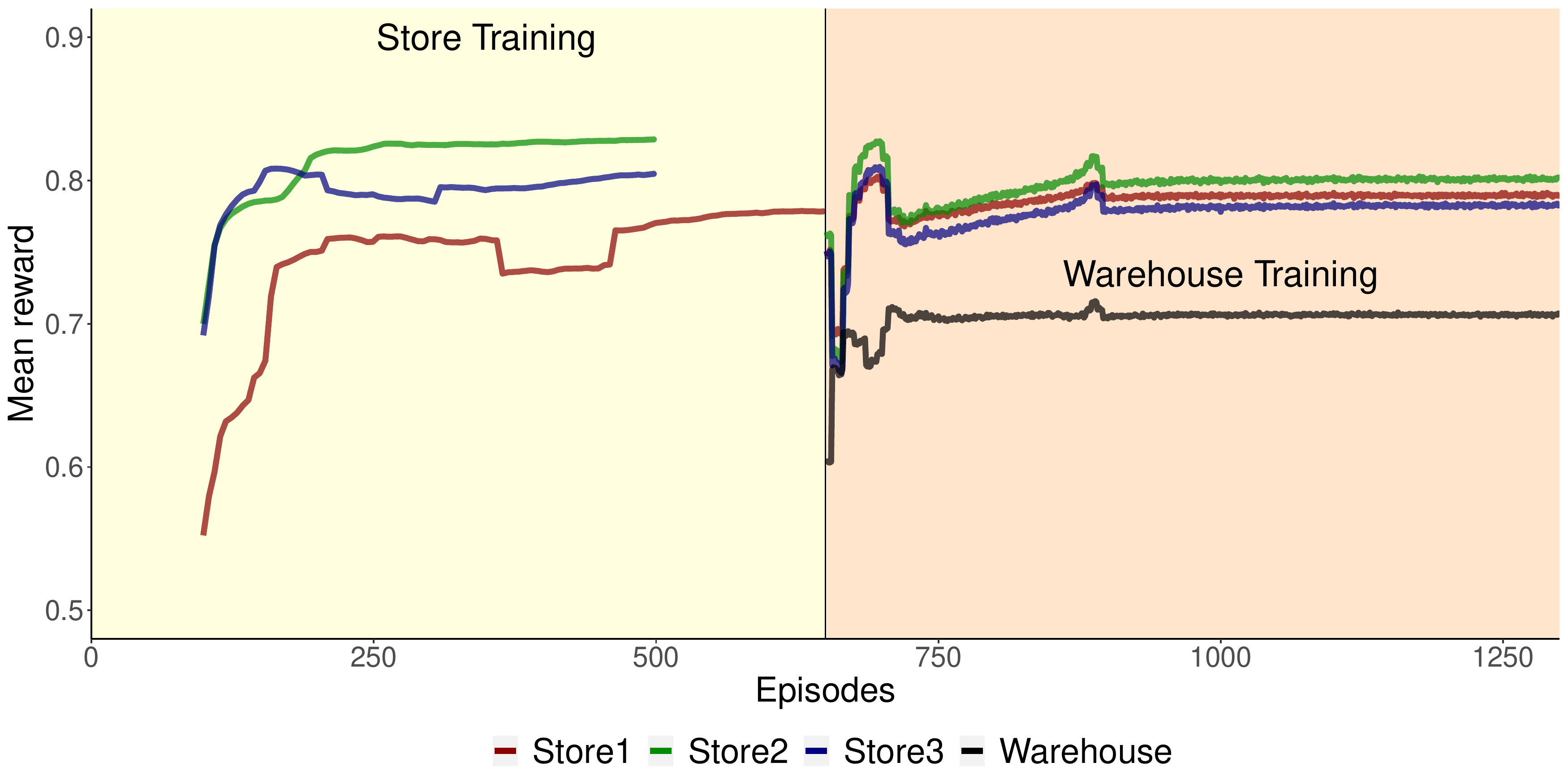}
  \caption{Training reward of Warehouse and Store for the 220 products}
  \label{fig:train_reward_220}
\end{figure}

\begin{figure}[h]
  \centering
  \includegraphics[width=0.9\textwidth]{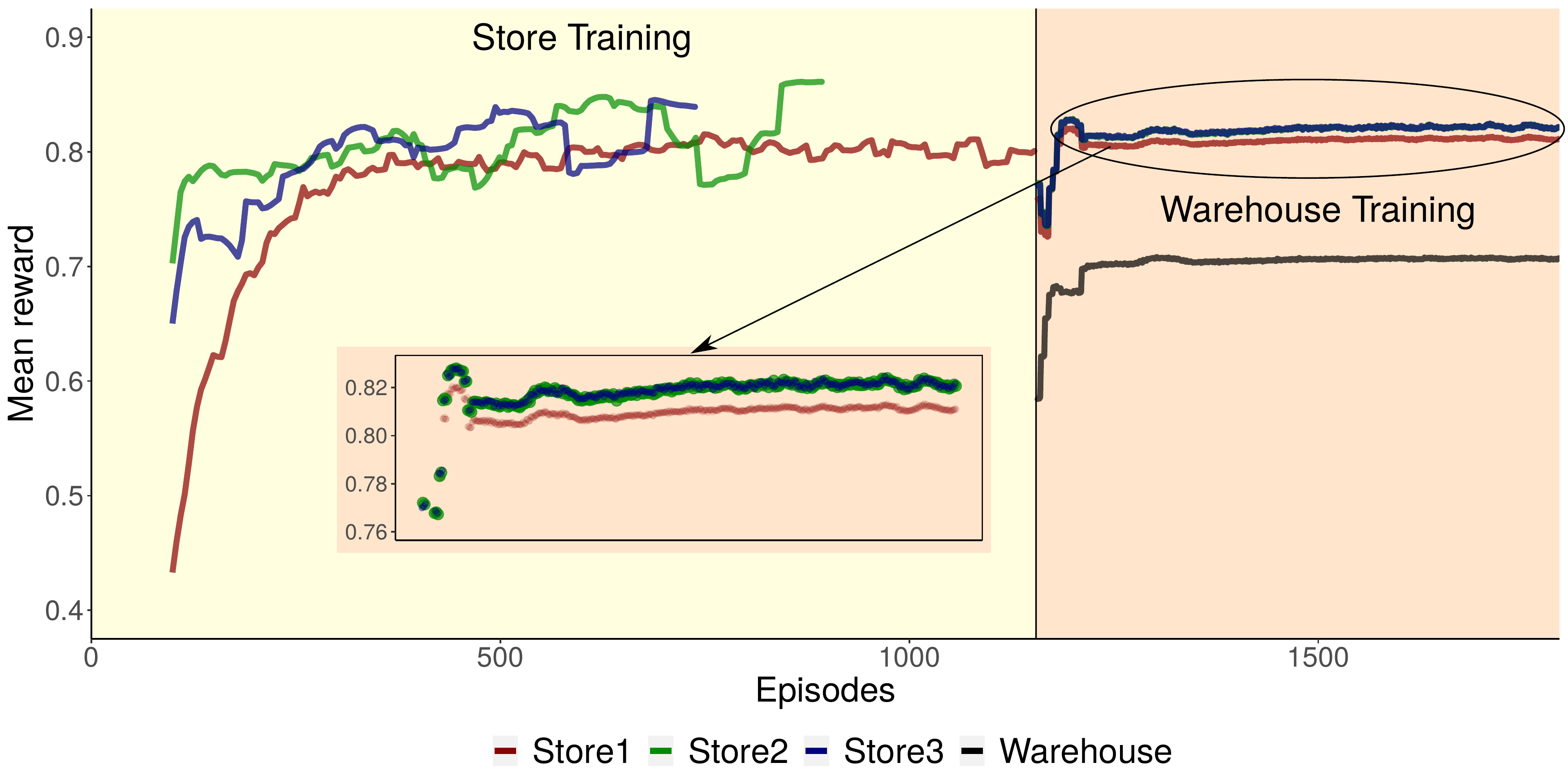}
  \caption{Training reward of Warehouse and Store for the 1000 products}
  \label{fig:train_reward_1000}
\end{figure}

\textbf{Training results: }
Fig.~\ref{fig:train_reward_50}, Fig.~\ref{fig:train_reward_220}, and Fig.~\ref{fig:train_reward_1000} show the results during training with the curriculum as described above, for 50, 220, and 1000 products respectively. In each plot, actions are being chosen using a multinomial distribution over the output probabilities. This allows us to stop training whenever a store or warehouse achieves convergence (variance of reward over last 50 episodes less than $10^{-4}$). When all three store agents achieve convergence, the warehouse training is initiated. All the plots also show variation of store rewards during warehouse training; we reiterate that these changes are due to variations in the warehouse behaviour, and the store policies themselves are fixed. Training steps are executed after every 5 episodes of experience collection (which can be parallelised), and the training uses a batch size of 128 time steps (including all products) and 40 epochs for each sample. Fig..~\ref{fig:train_reward_50}, Fig.~\ref{fig:train_reward_220}, and Fig.~\ref{fig:train_reward_1000} are segregated into two parts, where the first part depicts store training and the second part depicts warehouse training. The Y-axis is average reward across all time periods and products in an episode. We can see the rapid initial learning which is effectively achieved through multinomial sampling as explained earlier. For 1000 products it takes comparatively longer to achieve convergence which might be due to large number of effective actions making up the constraints. 

\textbf{Alternative policies for warehouse replenishment: }
Since the warehouse decisions are binary, we were able to experiment with several heuristic policies. Table~\ref{tab:result_50_diffpolicies} compiles these results. The policies that have not been introduced before are: (i) \textit{All ones}, where the warehouse replenishes all products on all days, (ii) \textit{All zeroes}, where no product is ever replenished in the warehouse, (iii) \textit{Alternate}, where each product is replenished in the warehouse in alternate time periods, and (iv) \textit{Random}, where each product has a 50\% chance of being replenished in any given time period. The results are shown for 50 products over training data. The efficiency of RL policy to effectively handle the trade-off between extremes is clearly visible. While \textit{All ones} achieves the best store rewards, it drastically reduces the warehouse rewards, effectively bringing down the system reward. RL is able to beat a purely random stochastic policy, emphasising that learning has taken place. The only policy that performs uniformly better than RL is Clairvoyant, which cheats by using true future sales values.

\begin{table}[h]
\caption{Different Warehouse policies with Stores on training data for 50 products}
\label{tab:result_50_diffpolicies}
\begin{center}
\begin{tabular}{|c|c|c|c|c|}
\hline
$\pi$ & Warehouse & Store1 & Store2 & Store3 \\ \hline
All ones & 0.126 & 0.811 &0.830 &0.832 \\ \hline
All zeros & 0.002 &-0.174 &-0.187 &-0.188 \\ \hline
Alternate & 0.346 &0.806 &0.234 &0.827 \\ \hline
Random & 0.355 &0.714 &0.732 &0.723 \\ \hline
RL & \textbf{0.464} &\textbf{0.798} &\textbf{0.816} &\textbf{0.812} \\ \hline
Heuristic & 0.280 &0.797 &0.803 &0.801 \\ \hline
Clairvoyant & 0.515 &0.810 &0.828 &0.829 \\ \hline
\end{tabular}
\end{center}
\end{table}

\begin{figure}
    \centering
    \subfigure[Wastage quantity]{\includegraphics[width=0.33\textwidth]{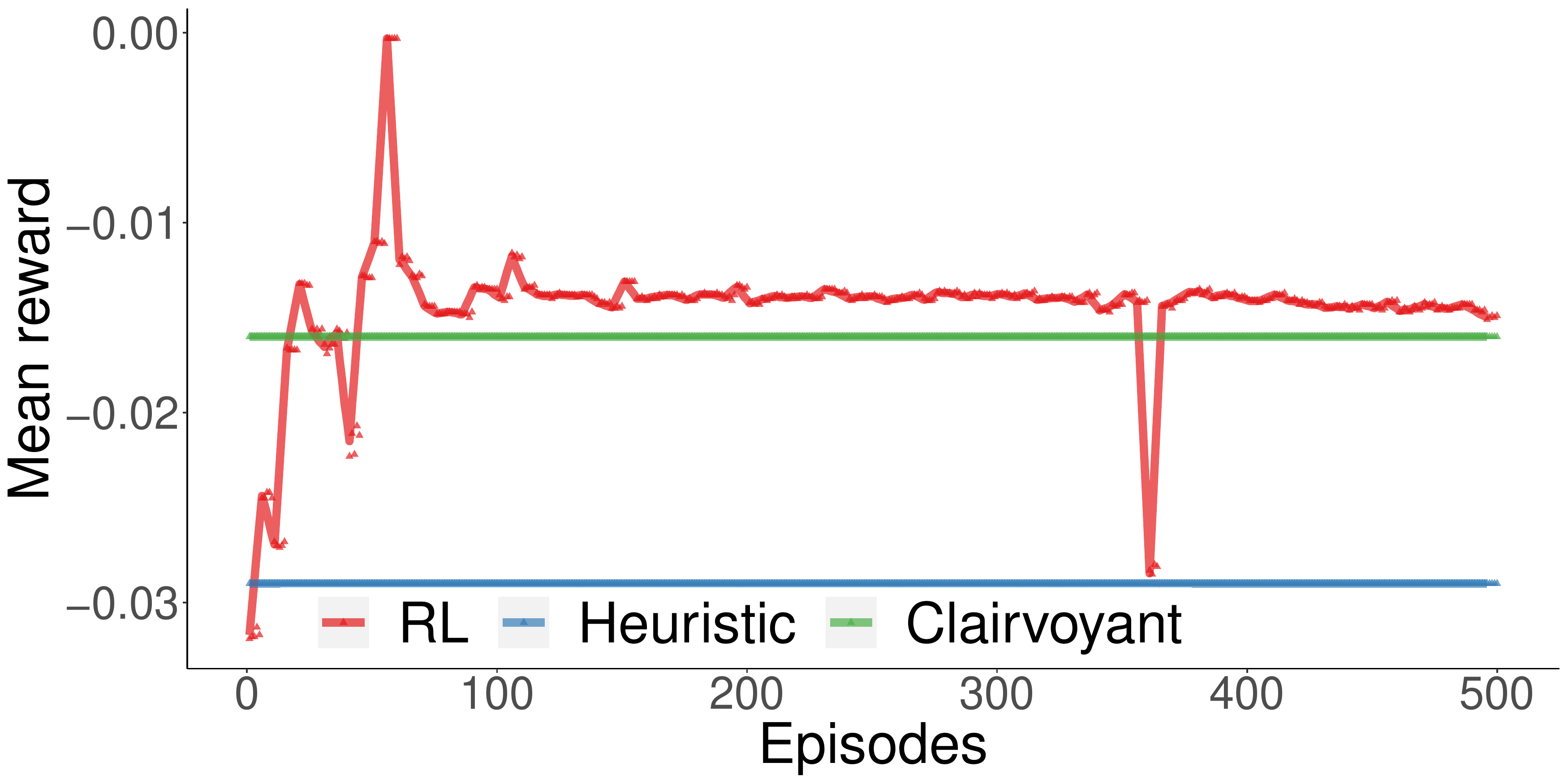}}
    \subfigure[Percentile reward]{\includegraphics[width=0.32\textwidth]{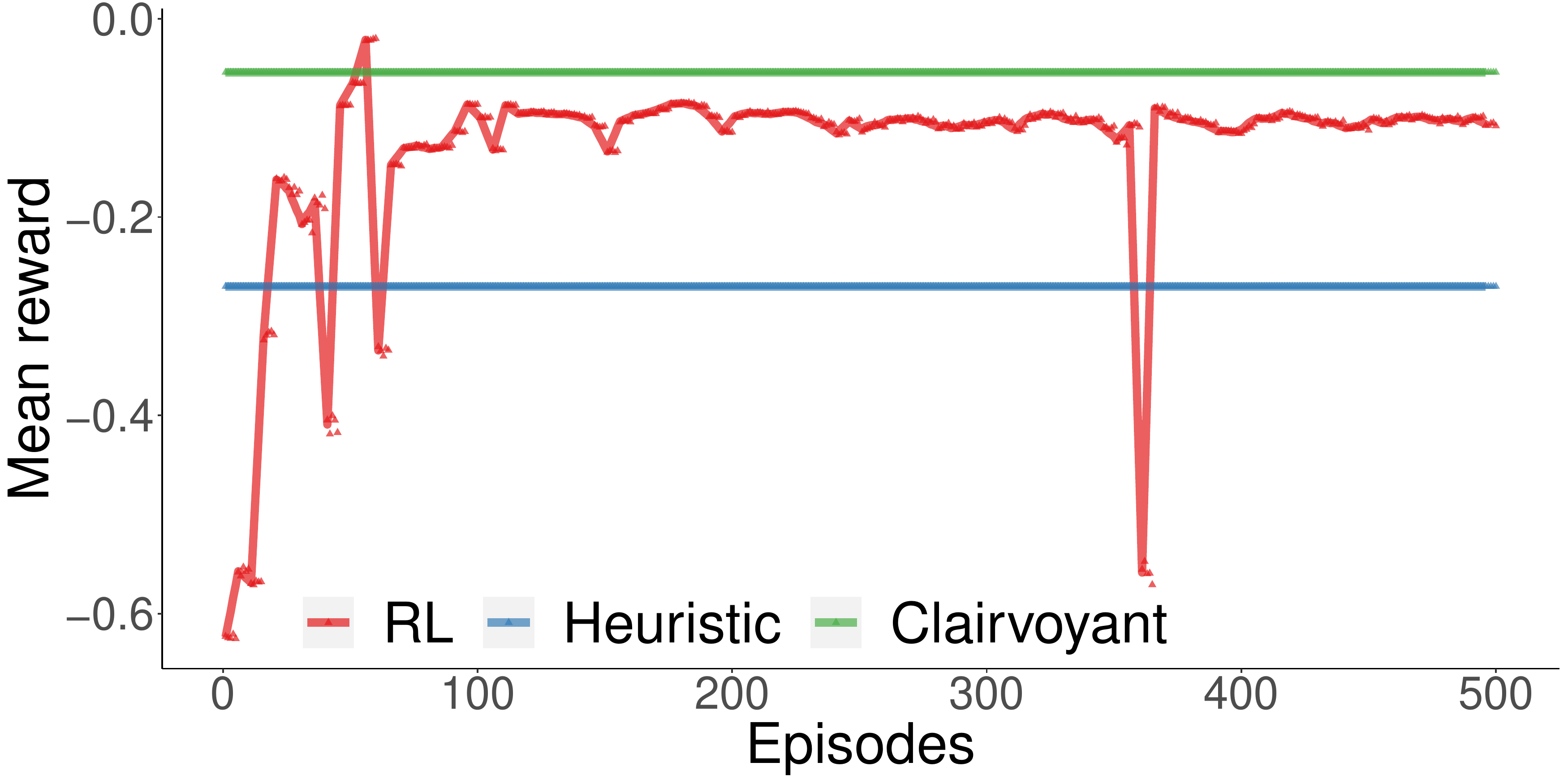}}
    \subfigure[Reward near zero]{\includegraphics[width=0.33\textwidth]{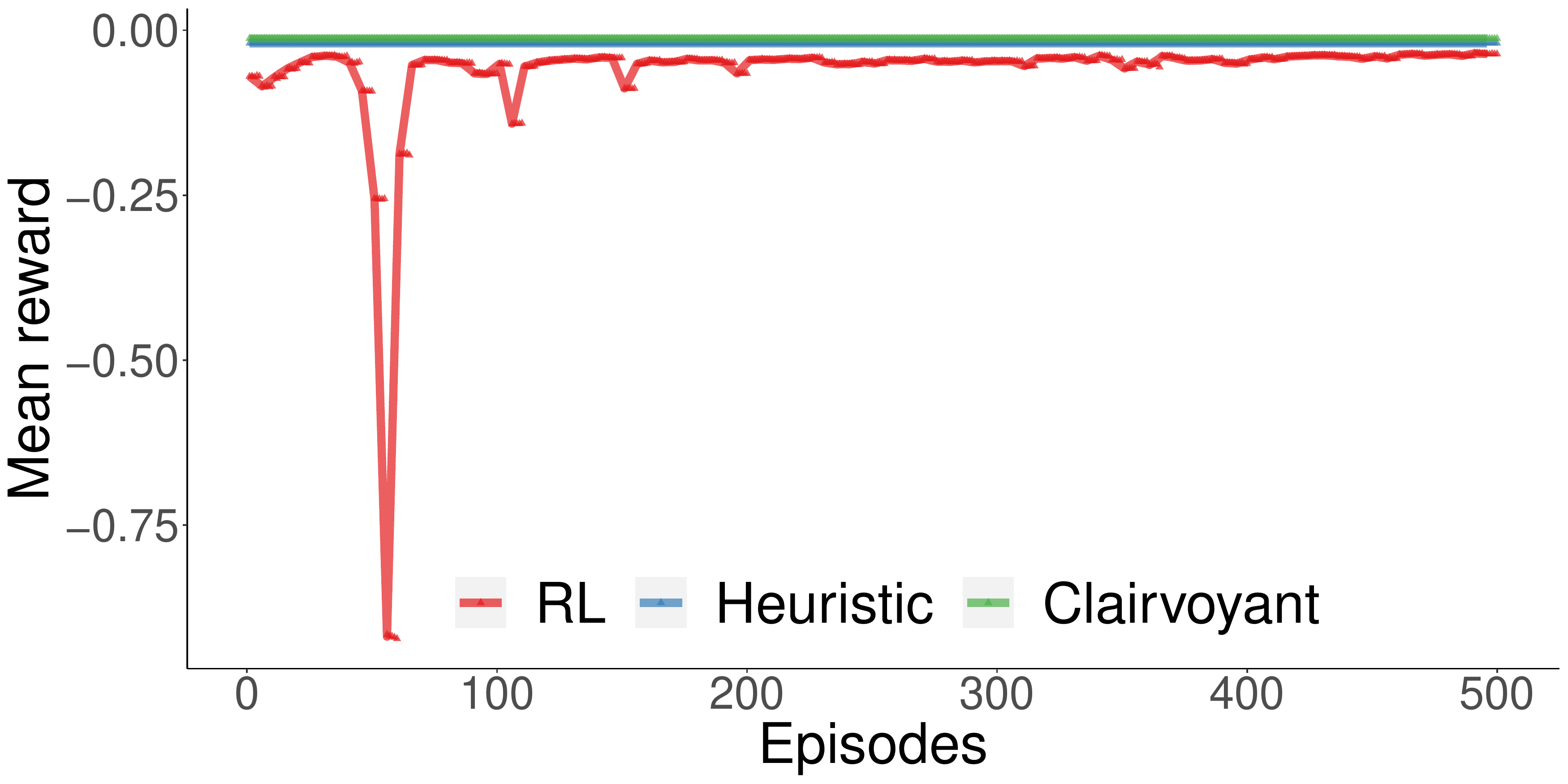}} 
    \subfigure[Refused order]{\includegraphics[width=0.33\textwidth]{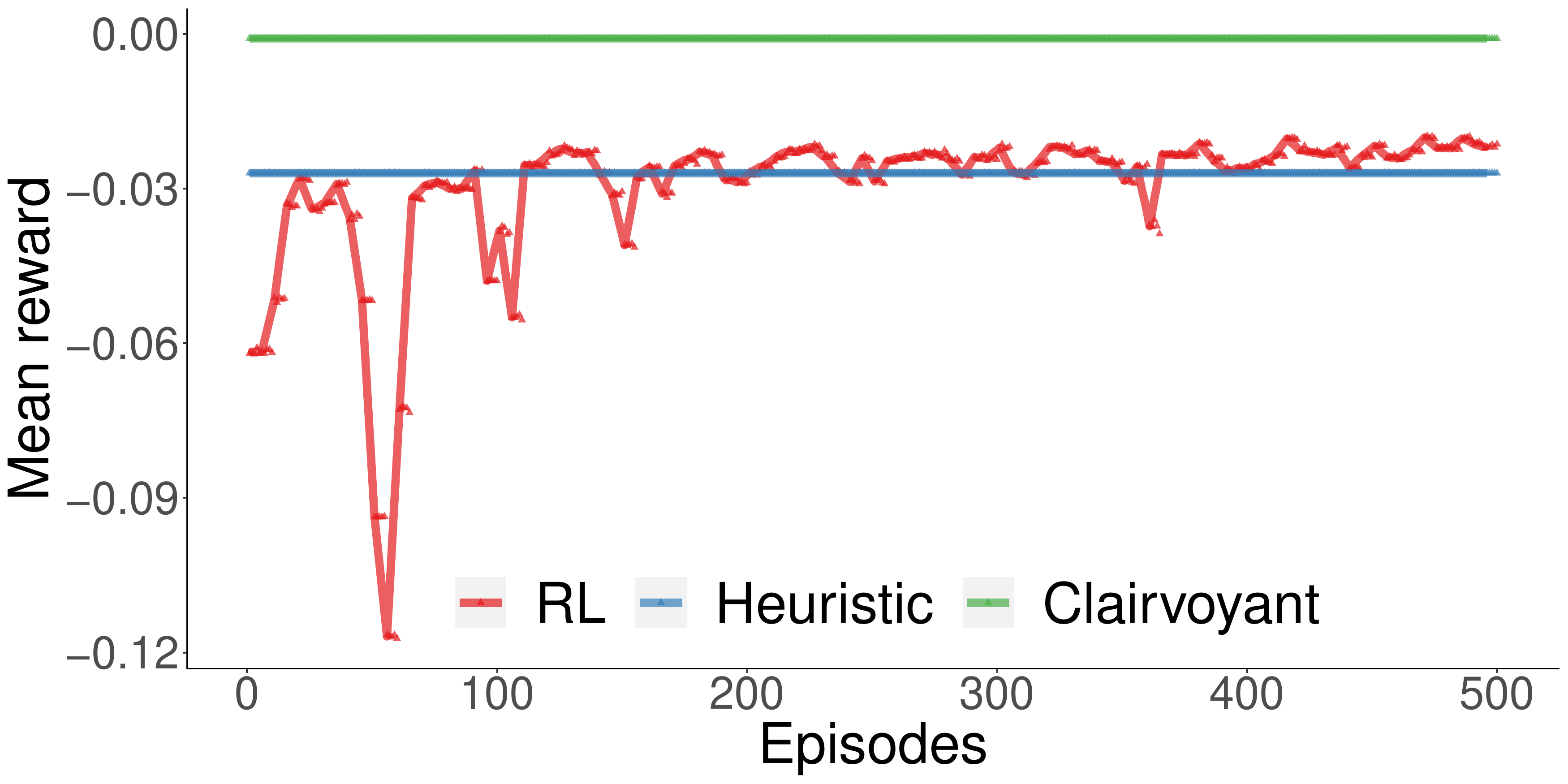}} 
    \subfigure[Average Inventory]{\includegraphics[width=0.32\textwidth]{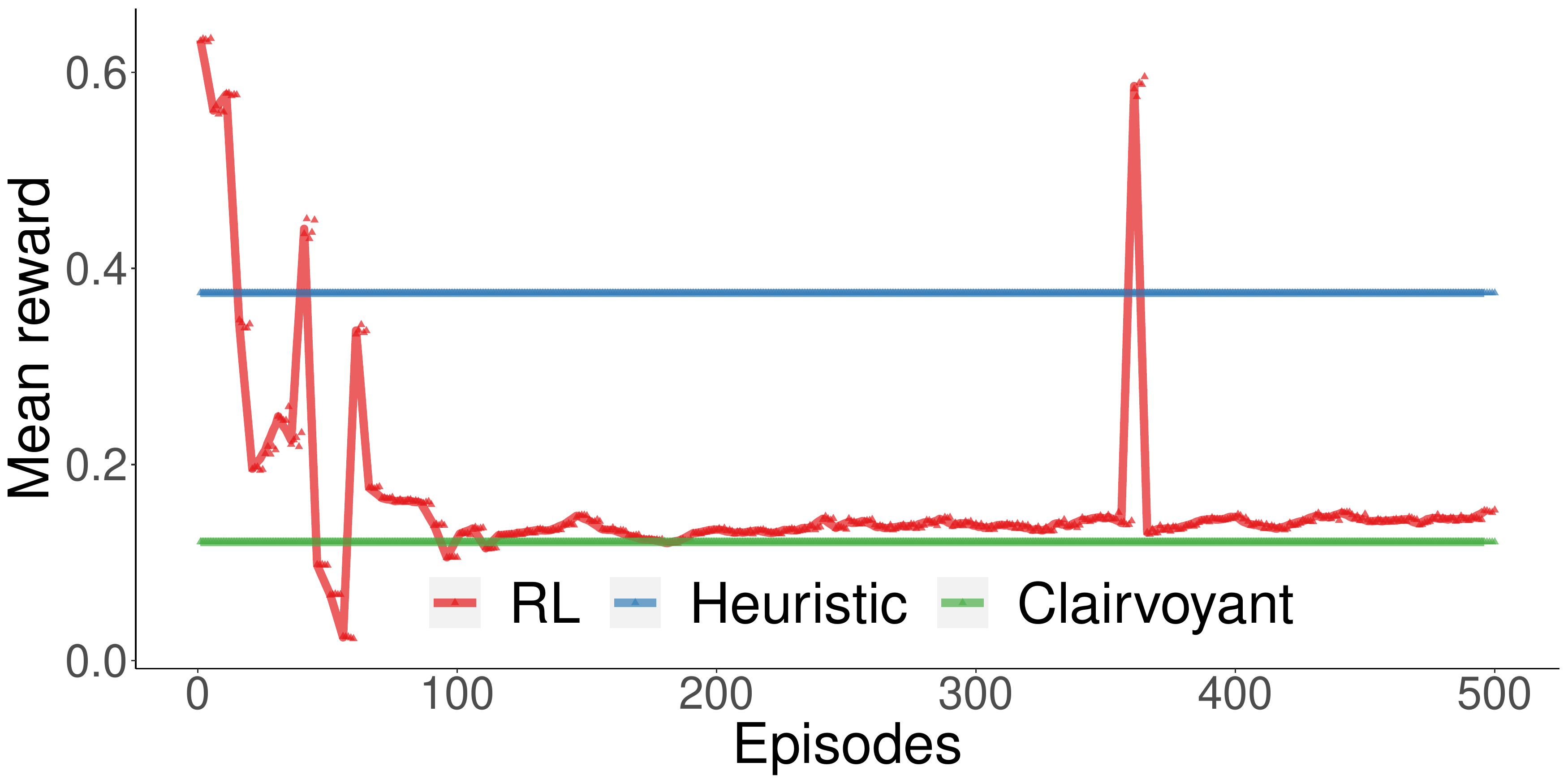}}
    \caption{Individual components of store replenishment rewards for Store 1 with 220 products.}
    \label{fig:store_220_IR}
\end{figure}
\begin{figure}
    \centering
    \subfigure[Wastage quantity]{\includegraphics[width=0.49\textwidth]{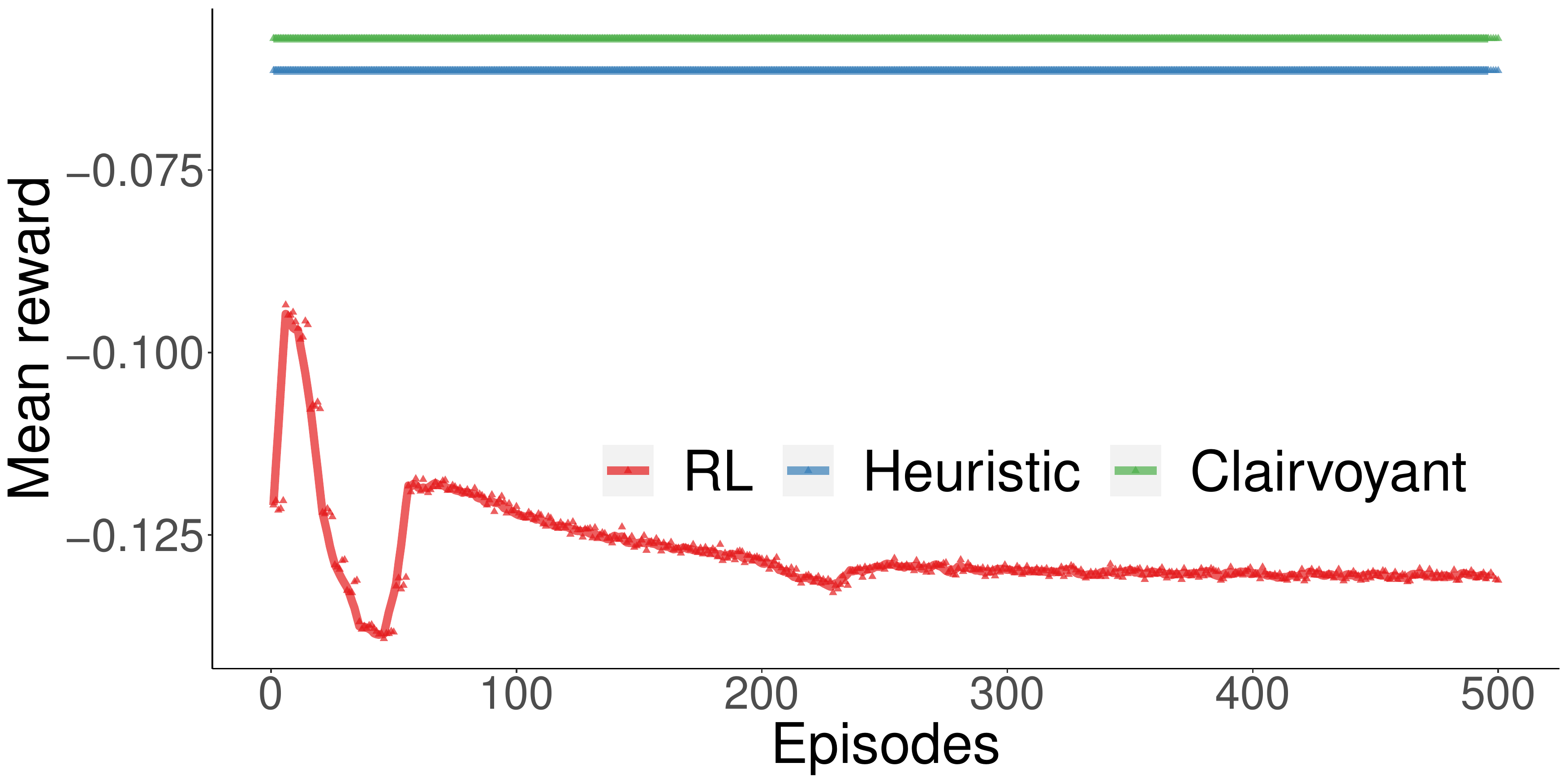}} 
    \subfigure[Store reward]{\includegraphics[width=0.49\textwidth]{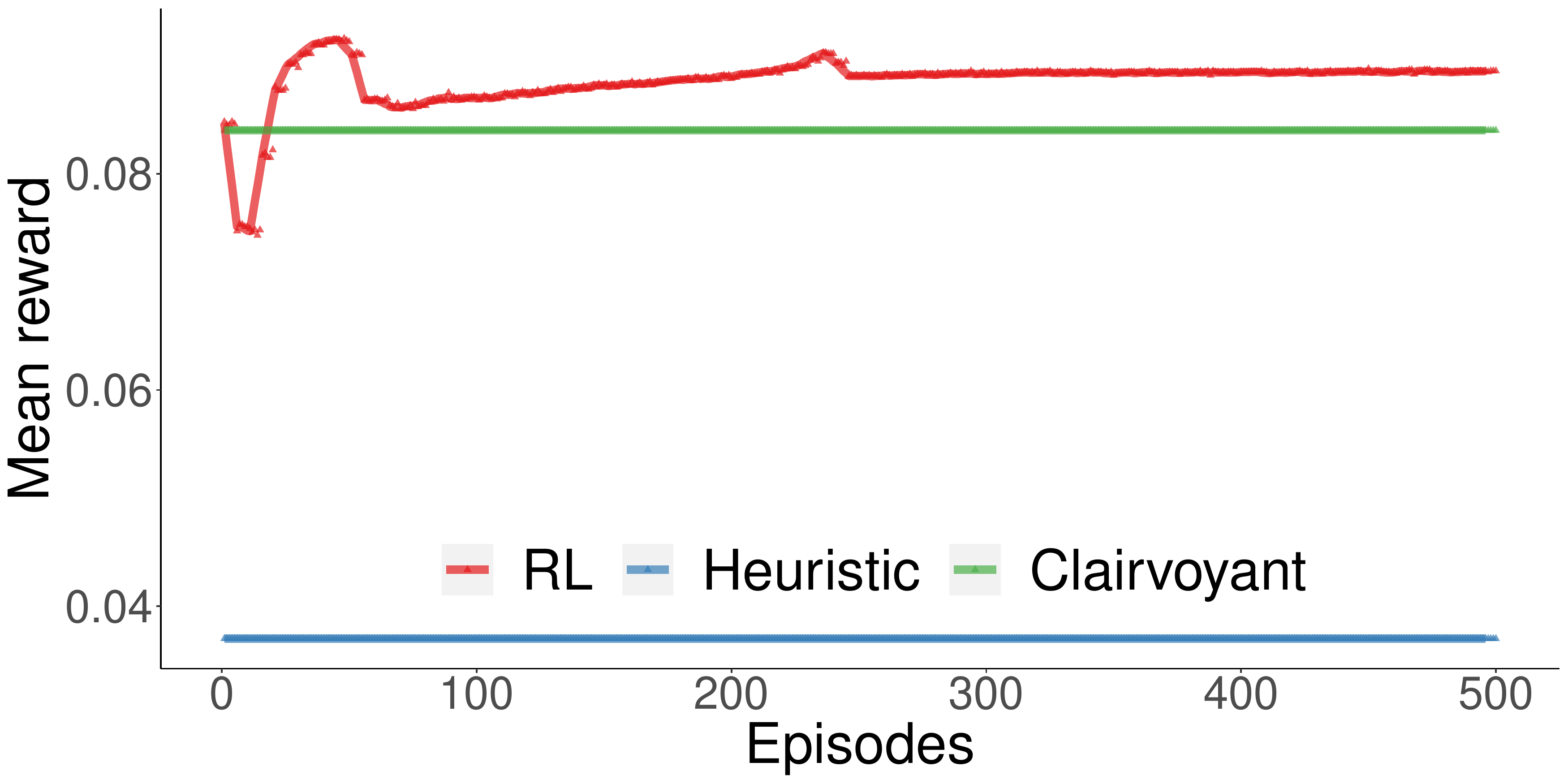}} 
    \subfigure[Replenish cost]{\includegraphics[width=0.49\textwidth]{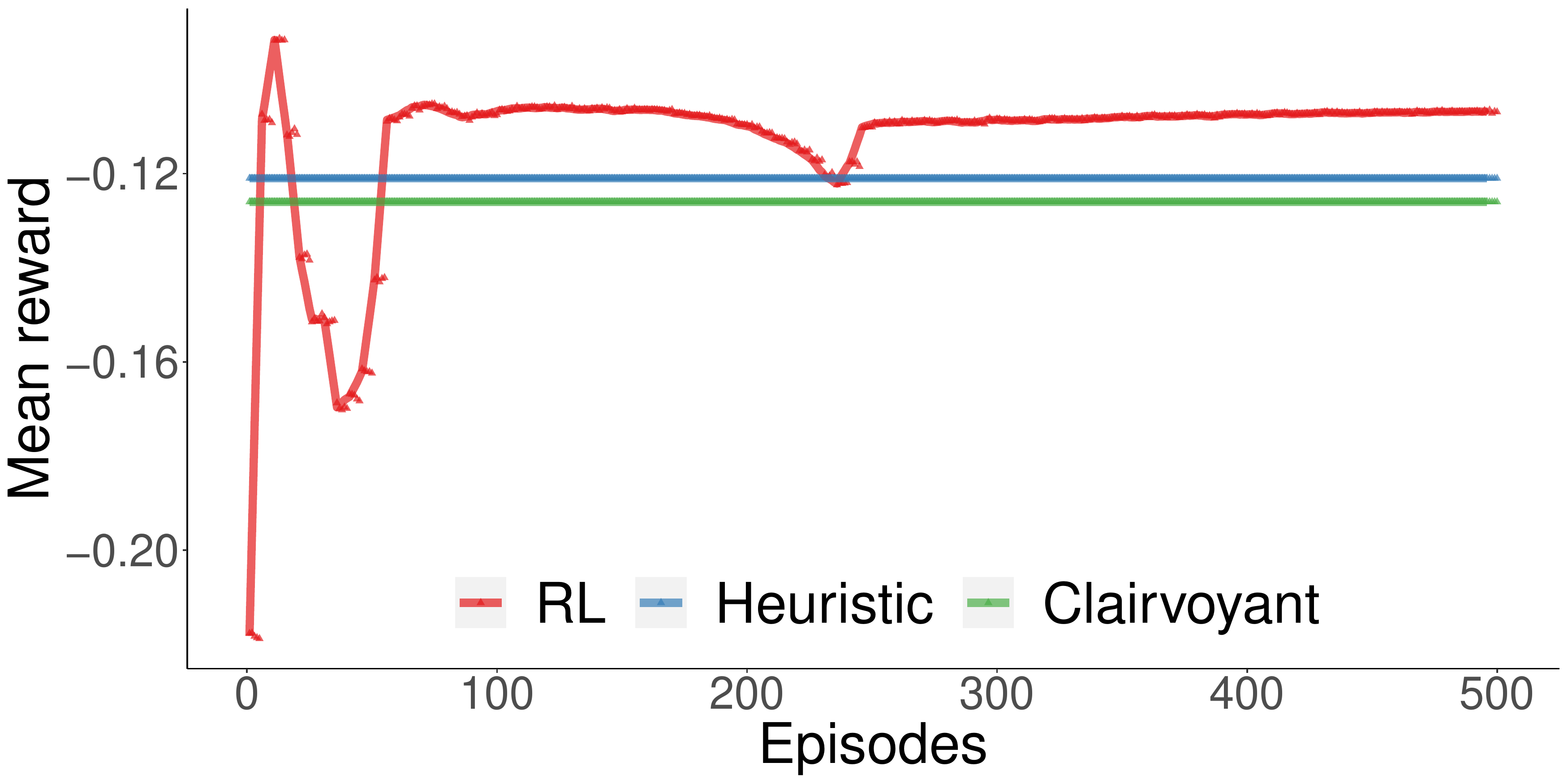}}
    \subfigure[Refused Orders]{\includegraphics[width=0.49\textwidth]{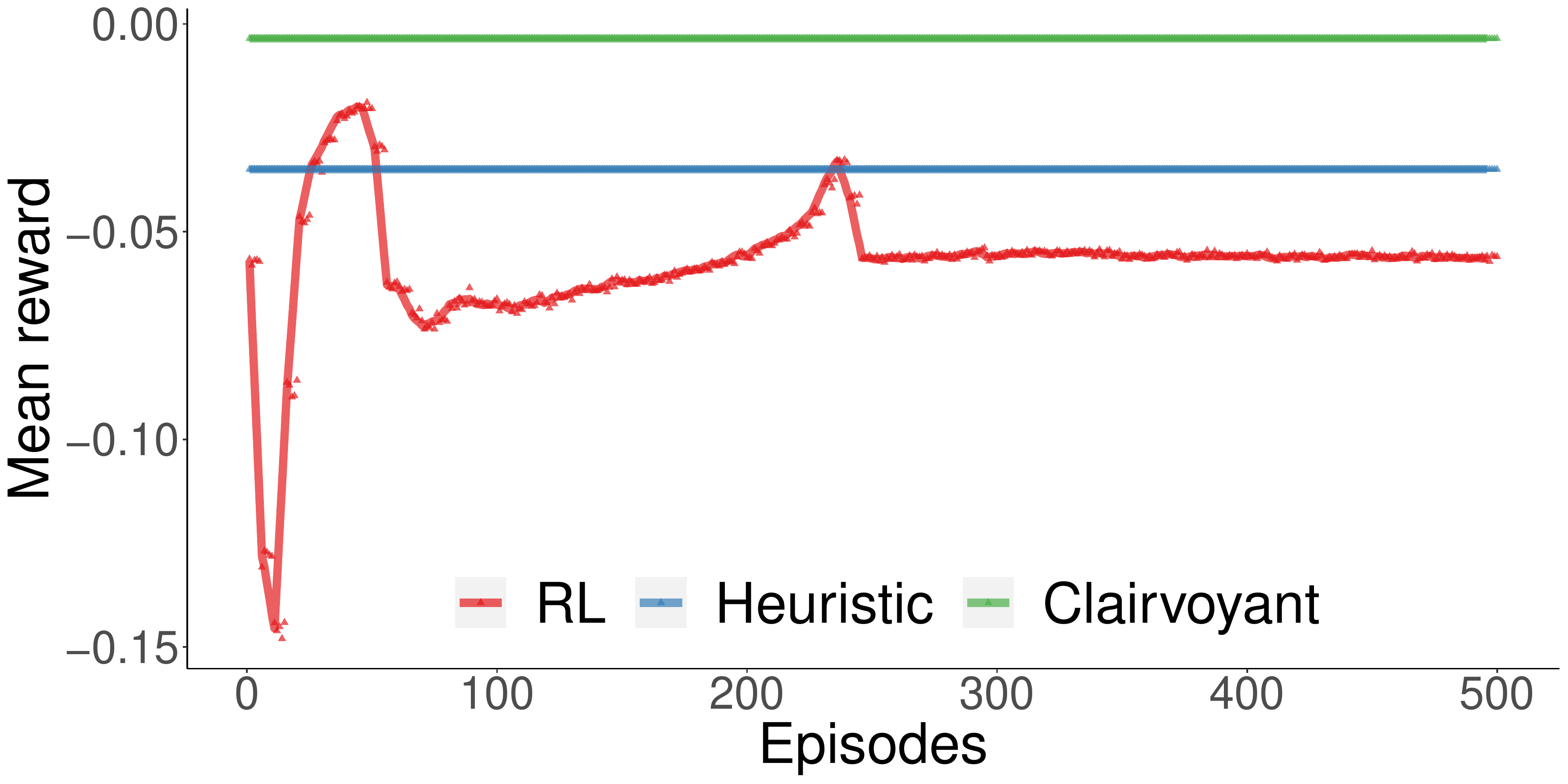}} 
    
    \caption{Individual components of warehouse rewards, for 220 products, 3 stores.}
    \label{fig:dc_IR}
\end{figure}

\textbf{Individual reward components: }
To understand the internal tradeoffs among the different components of total reward, we plot reward components of 220 product data and store 1 as shown in Figure~\ref{fig:store_220_IR}. We observe that the mean average reward across all the components of system reward decreases (negative magnitude) until convergence. Average inventory \ref{fig:store_220_IR}(e) for all stores decreases slightly to maintain the wastage and percentile reward, while keeping them above the stock-out level (near zero reward \ref{fig:store_220_IR}(c) is almost close to zero). Similarly for the warehouse, individual components of its system reward is shown in Figure \ref{fig:dc_IR}. The refused order in Figure \ref{fig:dc_IR}(d) and store reward in Figure \ref{fig:dc_IR}(b) are inversely proportional to each other since reduction in refusal from warehouse would improve the stores and vice versa. It learns to minimize the replenishment cost on warehouse while maintaining a balance between cost and wastage quantity.

\textbf{Results on test data: }
The results on an independent 4 months of test data (496 time periods) are compiled in Table \ref{tab:result_testdata}, where we observe that the overall performance using RL is better than the baseline constant-inventory heuristic algorithm on unseen data. Surprisingly, RL performs even better than the Clairvoyant policy for Store 1 in all three instances, and is also significantly better for Store 3 in the 1000 product instance. This is explained by the fact that the Clairvoyant policy - while it uses true future sales values - aims to maintain a user-defined constant inventory level. It is possible that depending on sales trends and forecasts, changing the inventory levels over time yields better rewards than keeping inventory fixed at all times.

\begin{table}
\centering
\caption{Results on testing data}
\label{tab:result_testdata}
\begin{tabular}{|l|l|l|l|l|}
\hline
Methods     & Warehouse & Store1 & Store2 & Store3 \\
\hline
            \multicolumn{5}{|c|}{50 products} \\
RL          & 0.451 & \textbf{0.793}  & 0.811  & 0.805 \\
Heuristic   & 0.392 & 0.617  & 0.671  & 0.690  \\
Clairvoyant & 0.552 & 0.766  & 0.838  & 0.817 \\
\hline
            \multicolumn{5}{|c|}{220 products} \\
RL          & 0.703 &    \textbf{0.783}     &   0.801      &  0.784 \\
Heuristic   & 0.738  & 0.610 & 0.695 & 0.692 \\
Clairvoyant & 0.769 & 0.716 & 0.839 & 0.836 \\
\hline
            \multicolumn{5}{|c|}{1000 products} \\
RL          & 0.702 & \textbf{0.796} & 0.820 &\textbf{0.818} \\
Heuristic   & 0.660 & 0.591 & 0.679 & 0.669 \\
Clairvoyant & 0.735 & 0.750 & 0.833 & 0.680 \\
\hline
\end{tabular}%
\end{table}

\textbf{Transfer learning: }
The key advantage of cloning the policy across products and execution in parallel is that the models can handle a change in the number of products without re-training\footnote{This is especially critical for real-world deployment. Retailers are constantly introducing new products and retiring new products, which means we cannot afford to build models that have to be retrained all the time.}. Table \ref{tab:transfer_learning_ondata} compares the results on a 70 product data set with the models trained on 50 products, versus the models trained from scratch on 70 products. The truck volume capacity has been proportionally scaled to be sufficient for 70 products. We observe that RL results for transfer learning and training from scratch have almost similar performance, and both models perform better than the heuristic.

For the warehouse, there is an additional dimension to transfer learning: instead of a change in the number of products, the warehouse also needs to handle a change in the number of stores being replenished. Table \ref{tab:transfer_learning_onstore} shows such an experiment, where the number of products remains the same, but `Store 4' is attached to the warehouse without retraining. The inputs to the warehouse model are rescaled (new normalisation factors) to accommodate the demand of the additional store, but no other changes are made. We see that RL is able to handle the new store seamlessly, and in fact performs better than the Clairvoyant policy. These results demonstrate the model’s capability to handle variable dimensions as well as additional nodes without disrupting any operational aspect of the inventory management problem. It thus forms the basis for a viable business solution.



\begin{table}[h]
\centering
\caption{Transfer learning of warehouse and stores on additional products.}
\label{tab:transfer_learning_ondata}
\begin{tabular}{|l|l|l|l|l|}
\hline
Methods     & Warehouse & Store1 & Store2 & Store3 \\
\hline
            \multicolumn{5}{|c|}{70 products} \\
\textbf{RL (transfer from 50 prod)}          & 0.428 & 0.764  & 0.811  & 0.796 \\
RL (trained on 70 prod)         & 0.431 & 0.784  & 0.819  & 0.812 \\
Heuristic   & 0.299 & 0.554  & 0.672  & 0.673  \\
Clairvoyant & 0.498 & 0.779  & 0.837  & 0.825 \\
\hline
\end{tabular}%
\end{table}

\begin{table}[h]
\centering
\caption{Transfer learning of warehouse on an unseen Store 4, with 1.75x capacity of Store 1.}
\label{tab:transfer_learning_onstore}
\begin{tabular}{|l|l|l|l|l|l|}
\hline
Methods     & Warehouse & Store1 & Store2 & Store3 & \textbf{Store4} \\
\hline
            \multicolumn{6}{|c|}{50 products} \\
RL          & 0.446 & 0.792  & 0.803  & 0.806 & 0.787 \\
Heuristic   & 0.373 & 0.619  & 0.672  & 0.693 & 0.651\\
Clairvoyant & 0.544 & 0.767  & 0.838  & 0.818 & 0.750 \\
\hline
\end{tabular}%
\end{table}



\begin{figure}
  \centering
  \includegraphics[width=\textwidth]{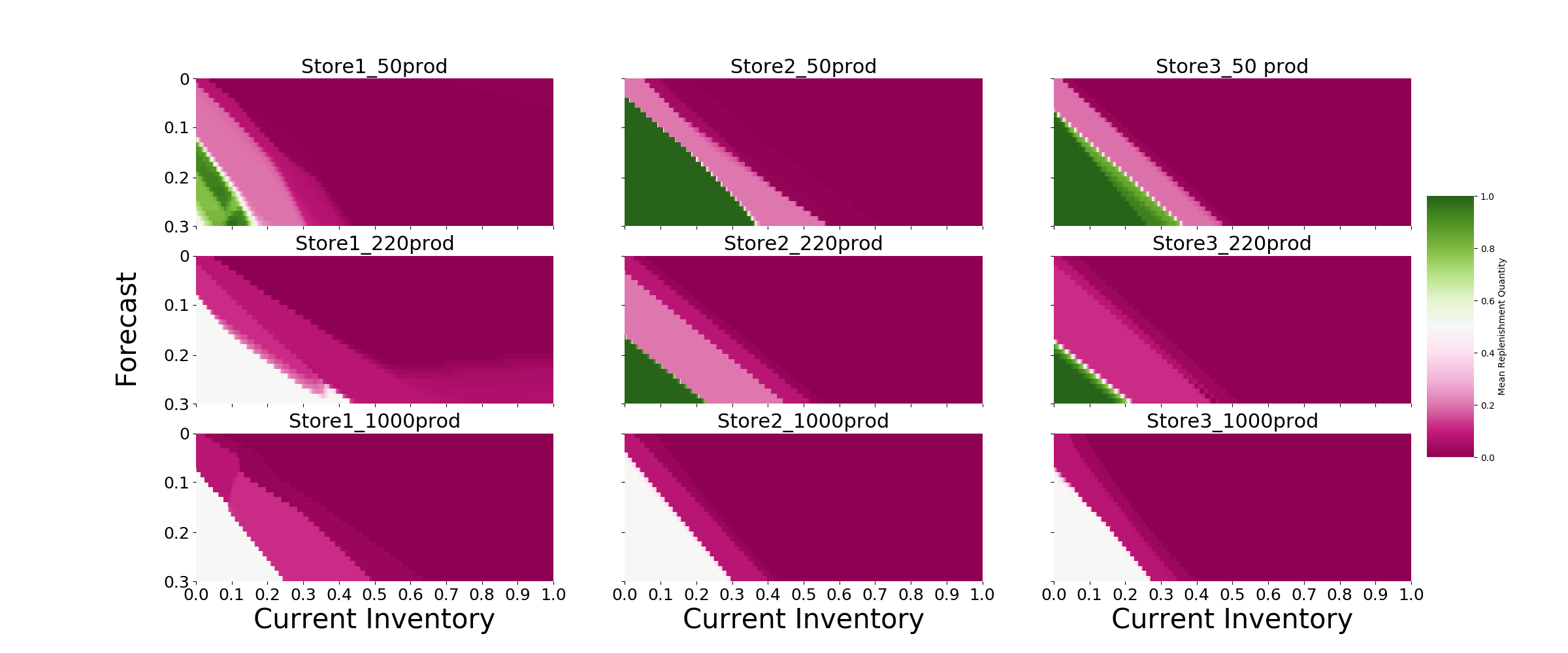}
  \caption{Policy heatmap for all stores and instances (50, 220, 1000 products) against current inventory and sales forecasts. Requested replenishment increases as current inventory decreases, and forecast increases (lower-left).}
  \label{fig:heatmap_allstore}
\end{figure}

\textbf{Policy insights: }
In order to understand the learned policies directly (rather than vicariously, through reward outcomes), we run experiments to measure the effect of different input features on the requested replenishment. We compare the requested replenishment action as a function of current inventory and forecast in Figure~\ref{fig:heatmap_allstore}. The figure shows a heatmap with the x-axis representing current inventory in the store and the y-axis representing corresponding forecast. We query the RL policy with all the permutation of the these two variables. In almost all the stores with low inventory and high forecast (lower-left corner) the replenishment actions are between 0.5 to 1 as normalized quantities. As inventory increases with lower forecast towards the top-right section the actions decreases to zero. However, when number of products are higher (1000 products), the RL policy asks for lower replenishment on average. This is possibly an effect of the law of large numbers; when the number of products is large, there is little chance of getting full replenishment (normalised output of 1.0) without incurring capacity exceedance penalties.

%

\section{Conclusion}
In this paper we described a reinforcement learning training paradigm for a real-world two echelon supply chain problem, illustrating the development with a single warehouse and three stores. A large varierty of experiments on three different problem instances (50, 220, 1000 products) were reported, as were transfer learning experiments for an unseen number of products (70 products) as well as stores (4 stores). All the different supply chain nodes were treated as separate RL agents and the training process was divided into two stages, where the first part comprises of store training (lower echelon) with infinite warehouse capacity, followed by warehouse training (higher echelon). The warehouse training includes the store performance into its scalar reward which greatly influences to take into consideration the whole system overall performance improvement. 

We wish to emphasise that this problem is very close to reality. Not only does it handle capacity constraints like finite warehouse and transportation capacities and a variable number of products, but is designed for deployment as an actual business solution. Our RL agents perform reasonably well, beating the heuristic on almost all occasions and performing close to its clairvoyant version, which suggests it is able to optimize performance in the long run. For future work, we are looking at an extension of the problem where each product has a different lead time for delivery from the suppliers to the warehouse. In conclusion, Reinforcement Learning could be a really useful strategy for solving such optimization problems and could be easily implemented practically providing better performance than current state-of-the-art strategies (mostly heuristics).


%
%

\bibliographystyle{spbasic}      
\bibliography{references}   

\end{document}